\pdfoutput=1

\documentclass[11pt]{article}

\usepackage[]{acl}

\usepackage{times}
\usepackage{latexsym}

\definecolor{snsblue}{HTML}{4c72b0}
\definecolor{snsorange}{HTML}{dd8452}
\definecolor{snsgreen}{HTML}{66c2a5}
\definecolor{snslightorange}{HTML}{fc8d62}
\usepackage[T1]{fontenc}

\usepackage[utf8]{inputenc}

\usepackage{microtype}
\usepackage{booktabs} 
\usepackage{multirow}
\usepackage{inconsolata}
\usepackage{colortbl}
\usepackage{xcolor}          
\usepackage{amssymb} 
\usepackage{subcaption}
\usepackage{subfloat}
\usepackage{graphicx}
 \usepackage{amsmath}
\newcommand\sect[1]{\S\ref{#1}}

\usepackage{tablefootnote}

\title{Amuro \& Char: Analyzing the Relationship between \\ Pre-Training and Fine-Tuning of Large Language Models}

 \author{Kaiser Sun \quad Mark Dredze \\
 Johns Hopkins University
 \\ Baltimore, MD USA \\
\texttt{\{hsun74,mdredze\}@cs.jhu.edu}\\
}

\begin{document}
\maketitle
\begin{abstract}

Large language model development relies on the pre-train-then-align paradigm, in which the model is typically pre-trained on a large text corpus and undergoes a tuning stage to align the model with human preference or downstream tasks.
We investigate the relationship between pre-training and supervised fine-tuning by considering multiple tasks as well as different pre-trained model checkpoints. Our results on 18 datasets and two models suggest that i) although the model benefits significantly through supervised fine-tuning, it may forget previously known domain knowledge and tasks that are not seen during fine-tuning; ii) the model exhibits high sensitivity to evaluation prompts after supervised fine-tuning, but this sensitivity can be alleviated through further pre-training; iii) continual pre-training improves the model in a latent way that manifests after fi
ne-tuning; iv) The model can already solve some tasks after pre-training, while fine-tuning most benefits datasets where the model does not show capability during pre-training.
\footnote{Code, results, and data to reproduce the experiments are available at \href{https://github.com/KaiserWhoLearns/AmuroCharPTFTRelationship}{github.com/KaiserWhoLearns/AmuroCharPTFTRelationship}. All the model checkpoints resulting from this work are available at 
\href{https://huggingface.co/KaiserWhoLearns/PTvsSFT_OLMo1b}{huggingface.co/KaiserWhoLearns/PTvsSFT\_OLMo1b}}

\end{abstract}

\section{Introduction}

The rise of large language models (LLMs) as a general-purpose tool for a diverse range of natural language processing tasks has dramatically transformed the field, introducing new paradigms for data collection and model training 
(\citealp{brown2020language}, \citealp{biderman2023pythia}, \citealp{touvron2023llama}, 
\citealp{jiang2023mistral}, 
\citealp{chowdhery2023palm}, \citealp{groeneveld2024olmo}, \citealp{wang2024helpsteer2}, \textit{inter alia}).
Numerous models, training methods, datasets, and evaluation methods continue to be developed on an ongoing basis.
Nevertheless, a unified paradigm has emerged for training LLMs: pre-train on an enormous corpus of diverse documents, ranging from 250B \citep{biderman2023pythia} to 15T \citep{llama3modelcard} tokens, followed by an alignment stage to make the model more useful and performative for various tasks.

Based on this paradigm, work has focused on improving these two stages. 
Work to improve pre-trained models includes larger training sets \citep{hoffmann2022training, llama3modelcard, touvron2023llama}, different data selection mechanisms \citep{xia2024less}, higher quality data \citep{zhou2024lima}, and various model architectures \citep{su2024roformer, touvron2023llama}. 
Meanwhile, research on model alignment includes different training objectives \citep{rafailov2024direct, schulman2017proximal},
new datasets \citep{narayanan-aepli-2024-tulu-resource}, more efficient training \citep{hu2021lora,dettmers2024qlora} and safety tuning \citep{bianchi2023safety}. The alignment stage usually involves either supervised fine-tuning for specific tasks or instruction fine-tuning for general-purpose usage. 
Regardless, fine-tuning (almost always) comes at the end of pre-training and yields remarkable improvements on downstream tasks \citep{touvron2023llama, groeneveld2024olmo}. 
\begin{figure}[t]
  \includegraphics[width=\columnwidth]{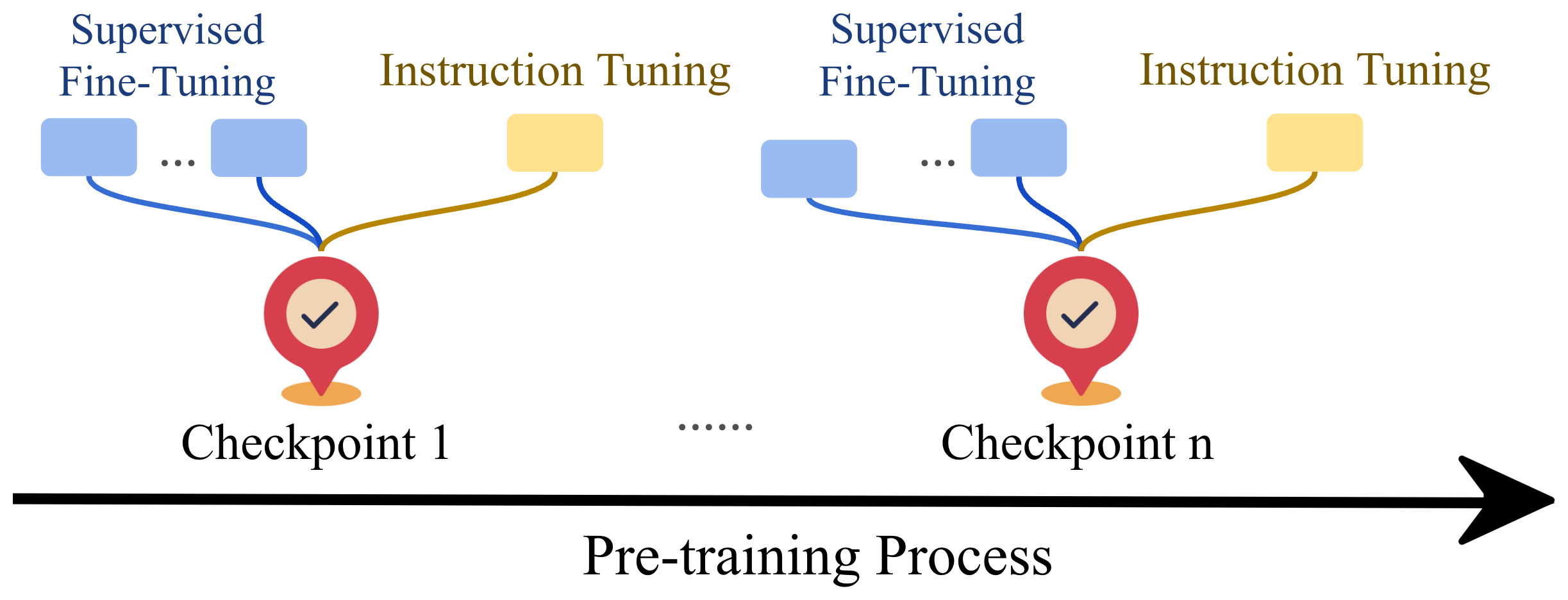}
  \caption{Illustration of the experimental scheme. Intermediate pre-training checkpoints are fine-tuned on different datasets.}
  \label{fig:experiment-illu}
  \vspace{-12pt}
\end{figure}

Consequently, the benefits of each stage are largely explored independently, with improvements to pretraining being orthogonal to benefits from model alignment.

Rather than exploring these two training regimes independently, we ask: What does the model learn and forget during pre-training and fine-tuning? Specifically, {\bf how do pretraining and fine-tuning interact to produce the resulting model?} Does more pre-training hinder better fine-tuning results? 
Answering these questions requires us to examine how models learn during pre-training and how this affects fine-tuning. 
Therefore, we begin by fine-tuning two language models under a variety of conditions to determine how fine-tuning affects model behavior.
We explore both supervised and instruction fine-tuning, testing the models' memorization and forgetting when learning specific tasks and serving as general-purpose language-AI tools.
We then explore the affect of pre-training on these behaviors by fine-tuning {\bf multiple pre-training checkpoints} of a large language model (Figure~\ref{fig:experiment-illu}), evaluating each checkpoint and its fine-tuned variant on downstream evaluation sets.
We track model abilities during pre-training and compare them to improvements achieved after fine-tuning at the corresponding pre-training step.\footnote{While we believe that we were the first to explore these issues through intermediate model checkpoints, recently released work has also utilized pre-training checkpoints and are highlighted in Section~\ref{sec:relatedwork}.}

Our experiments yield the following insights into LLM training:
 (1) although supervised fine-tuning can improve performance on in-distribution tasks, it can also cause the model to forget domain knowledge or tasks that it was previously capable of solving (\sect{sec:finding:what}); (2) fine-tuned models show high sensitivity to evaluation prompts, but this sensitivity can be alleviated by more pre-training (\sect{sec:finding:what}); (3) continued pre-training can improve a model in ways that are only revealed after fine-tuning (\sect{sec:finding:PTFT}); (4) tasks for which the model already performs well during pre-training benefit much less from fine-tuning than those where the model does not demonstrate capabilities (\sect{sec:finding:base-eval}, \sect{sec:finding:PTFT}); 

Our findings provide insights into model training and can inform methods for both pre-training and fine-tuning. Furthermore, our work shows the value of analyzing the training dynamics, in addition to analyzing the final checkpoint of an LLM, as an aspect of interpretability, and we encourage model developers to release these checkpoints to aid future studies.

\section{Background: Model Training}

We use ``model alignment'' as a general term for techniques that align a model with a desired behavior,
often accomplished by fine-tuning models after pretraining. 
The term is also associated with other definitions \citep{shen2024bidirectional}.

We begin with a brief survey of the core components of LLM training: pre-training, fine-tuning, and instruction fine-tuning.
The first step of training an LLM is pre-training on a massive text corpus
 \citep{achiam2023gpt, touvron2023llama, groeneveld2024olmo}.
 Initial work increased model size to hundreds of billions of parameters \citep{brown2020language, rae2021scaling, chowdhery2023palm}, along with explorations in model size, training corpus size, and training data characteristics \citep{radford2019language, hoffmann2022training, gururangan-etal-2020-dont}. 
Other work increased the amount of pre-training data \citep{together2023redpajama, soldaini2024dolma}, with new models now reaching 15 trillion tokens \citep{llama3modelcard}.

After the pre-training stage, when a specific task of interest has been identified, supervised fine-tuning can improve a pre-trained model. 
Task-agnostic tuning became popularized with the advent of T5 models \citep{raffel2020exploring}, where a pre-trained LLM is tuned using a general text-to-text solution. 
Instruction fine-tuning is preferred when more general model behaviors are desired. 
When multiple tasks are given to the model, the model is commonly given a task-specific prefix or an instruction along with the task input, leading to the development of various methods of prefix tuning \citep{li-liang-2021-prefix} and instruction tuning \citep{wei2021finetuned, mishra-etal-2022-cross, victor2022multitask}.

Other works explore human preference tuning with or without a reward model \citep{christiano2017deep, ziegler2019fine, stiennon2020learning, ouyang2022training, rafailov2024direct, song2024preference, xu2024contrastive}.
In-context learning utilizes a small amount of supervised data to improve model performance without updating the parameters.
In this work, we focus specifically on single-task supervised fine-tuning and multi-task instruction tuning.

\section{Experimental Setup}
In this section, we describe the models and datasets used. 
The hyperparameter tuning procedure and setup for each fine-tuning setting can be found in Appendix~\ref{sec:app:hyperparameter}.

\begin{table}[t!]
\centering
\small
\begin{tabular}{@{}llll@{}}
\toprule
\multicolumn{4}{c}{\textbf{Supervised Fine-Tuning}} \\ \midrule
\textbf{Task} & \textbf{Training} & \textbf{ID Test} & \textbf{OOD Test} \\ \midrule
\begin{tabular}[c]{@{}l@{}}Summary \\ Generation\end{tabular} & XSum & \begin{tabular}[c]{@{}l@{}}XSum, \\ XLSum\end{tabular} & CNN \\  \arrayrulecolor{black!30}\midrule
\begin{tabular}[c]{@{}l@{}}Question \\ Generation\end{tabular} & SocialIQa & SocialIQA & \begin{tabular}[c]{@{}l@{}}SciQ, \\ TweetQA\end{tabular} \\  \arrayrulecolor{black!30}\midrule
\begin{tabular}[c]{@{}l@{}}Natural Language \\ Inference\end{tabular} & MNLI & \begin{tabular}[c]{@{}l@{}}MNLI1, \\ MNLI2\end{tabular} & \begin{tabular}[c]{@{}l@{}}RTE, \\ GPT3NLI\tablefootnote{\href{https://huggingface.co/datasets/pietrolesci/gpt3_nli}{https://huggingface.co/datasets/pietrolesci/gpt3\_nli}} \end{tabular} \\  \arrayrulecolor{black!30}\midrule
\begin{tabular}[c]{@{}l@{}}Paraphrase \\ Detection\end{tabular} & Paws & Paws & \begin{tabular}[c]{@{}l@{}}QQP, \\ STS-B\end{tabular} \\ \arrayrulecolor{black}\midrule

\multicolumn{4}{c}{\textbf{Instruction Tuning}} \\ \midrule
\textbf{Dataset} & \multicolumn{3}{l}{\textbf{Description}} \\
\midrule
T\"ULU-v2 & \multicolumn{3}{l}{A mixture of instruction datasets.} \\

ARC & \multicolumn{3}{l}{Grade-school multiple-choice QA.}   \\
OpenbookQA &  \multicolumn{3}{l}{Open book exam QA.} \\
Hellaswag &  \multicolumn{3}{l}{Commonsense inference.} \\
BoolQ & \multicolumn{3}{l}{Reading comprehension.}  \\
SciQ & \multicolumn{3}{l}{Science exam multiple choice QA.}  \\ \arrayrulecolor{black}\bottomrule

\end{tabular}
\caption{Dataset information. For Generation tasks, ROUGE-L is used as evaluation metric, and accuracy is used for classification tasks. ID = In-domain, OOD = Out-of-domain.}
\label{tab:tasks}
\vspace{-10pt}
\end{table}

\subsection{Model Choice} 
We consider two open models of different architectures and scales: Llama3-8B \citep{llama3modelcard} and OLMo-1B \citep{groeneveld2024olmo}.
To minimize potential confounding factors such as multilingual ability and double descent \citep{belkin2019reconciling, caballero2022broken, schaeffer2023double}, we exclusively select models predominantly pre-trained in English and incorporate significantly more pre-trained tokens than the number of parameters.
We do not include models trained in a multi-stage manner to ensure uniformity of the tokens seen by the model during pre-training.
Some of our experiments consider intermediate pre-training checkpoints.
We select checkpoints uniformly by the number of tokens seen from the pre-training history along with the first and the final checkpoints.
Unfortunately, very few large language models release intermediate pre-training checkpoints (summarized in Table~\ref{tab-app-checkpoints}).
Further consideration and reasoning of model selection are included in Appendix~\ref{sec:app:modelselection}.

\subsection{Training Procedure}
We fully fine-tune each of the selected model checkpoints using two different procedures to create fine-tuned models: supervised fine-tuning and instruction tuning. 
The supervised fine-tuning is conducted separately for each model checkpoint and dataset, while the instruction fine-tuning is done once using the instruction dataset.
The instruction-tuned model is evaluated on a suite of LLM benchmarks.
All experiments are conducted on two Nvidia 80GB A100, with a total cost of approximately 1100 GPU hours.
The detailed number of GPU hours consumed for each experiment is included in Appendix~\ref{sec:app:gpuhours}.

\textbf{Supervised Fine-tuning. }
We adapt the datasets from \citet{yang2024unveiling} for supervised fine-tuning. 
For each in-domain dataset, one to two cross-domain evaluation datasets are supplied.
OLMo-1B is fully fine-tuned for 3 epochs with a batch size of 8, while Llama3-8B is fine-tuned with a batch size of 16 and 2 training epochs.
Both models are trained with learning rates resulting from minimal hyperparameter tuning (Appendix~\ref{sec:app:hyperparameter}).
Each task is formatted using a default prompt-completion format (Table~\ref{tab:app:promptformat}).

\textbf{Instruction Fine-Tuning. }
We instruction-tune the model on T\"{U}LU \citep{ivison2023camels}, following the decision of \citealp{groeneveld2024olmo}.
Each model checkpoint is fully fine-tuned for 5 epochs with a batch size of 8 and a learning rate of $2\times 10^{-6}$.

\subsection{Evaluation}
For each model, we conduct a few-shot evaluation with a shot size of 4, after examining with shot size in \{0, 2, 4, 6\}.

\begin{figure*}[t!]
    \centering
    \begin{subfigure}[b]{0.36\textwidth}
    \includegraphics[width=\the\columnwidth]{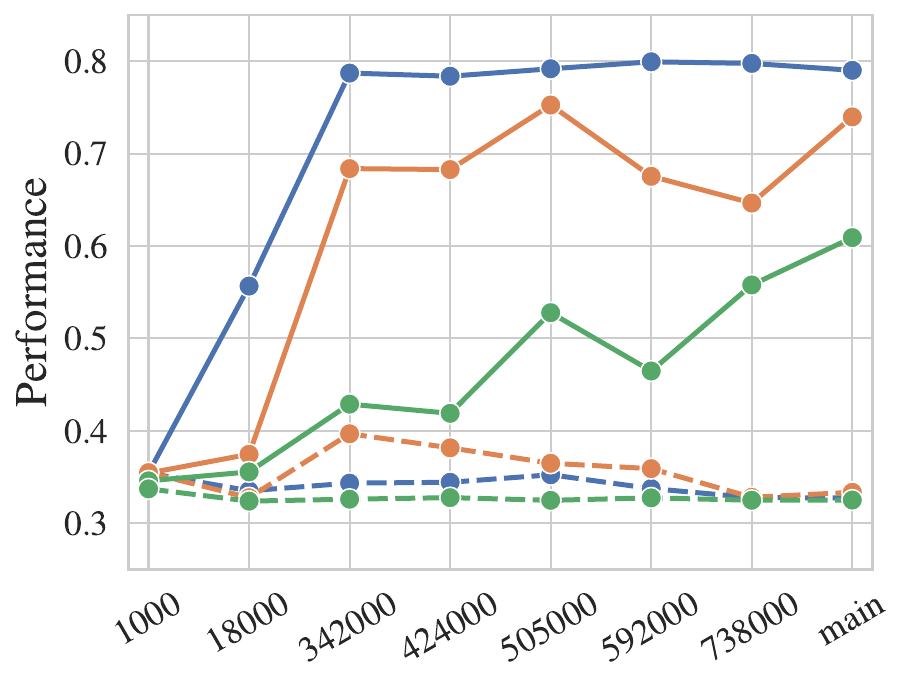}
        \caption{MNLI matched}
    \end{subfigure}%
    ~
    \begin{subfigure}[b]{0.51\textwidth}
    \includegraphics[width=\the\columnwidth]{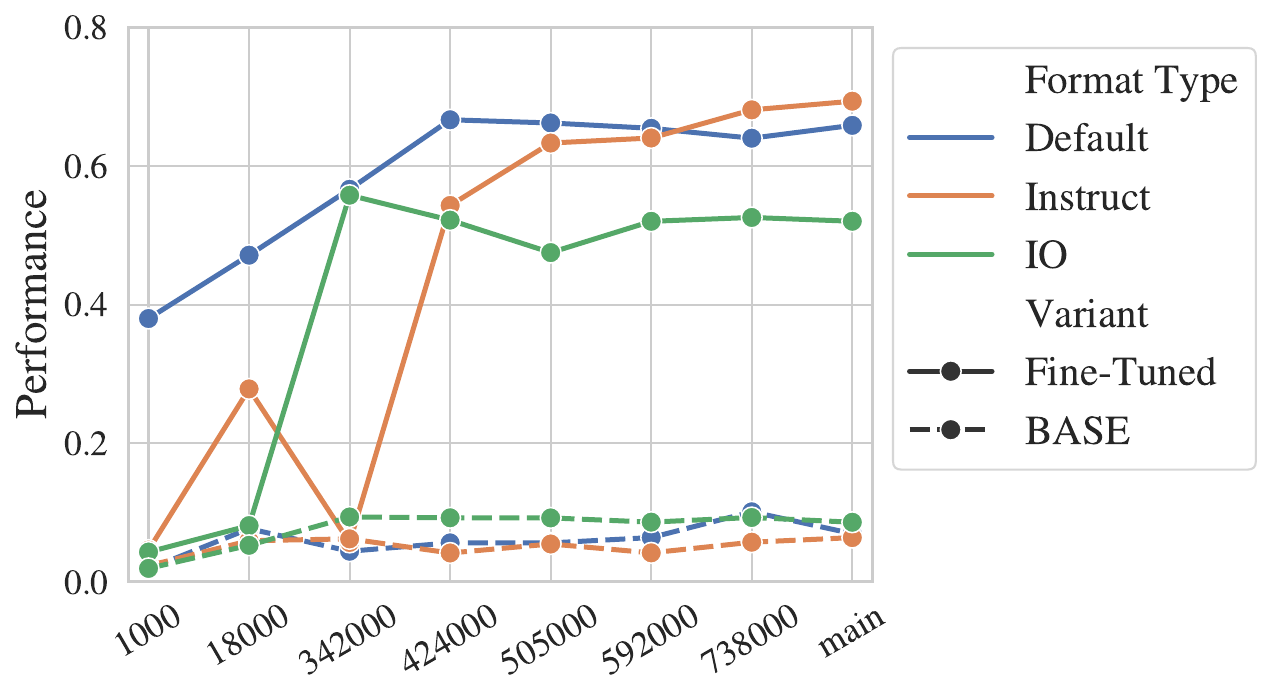}
        \caption{SocialIQa}
    \end{subfigure}%

    \caption {Example of model performance with different task formats. The figure of all datasets can be found in Figure~\ref{fig:app:task_format}. }
  \label{fig:findings:task_format}
  \vspace{-10pt}
\end{figure*}

\begin{figure}[t]
\centering
  \includegraphics[width=\columnwidth]{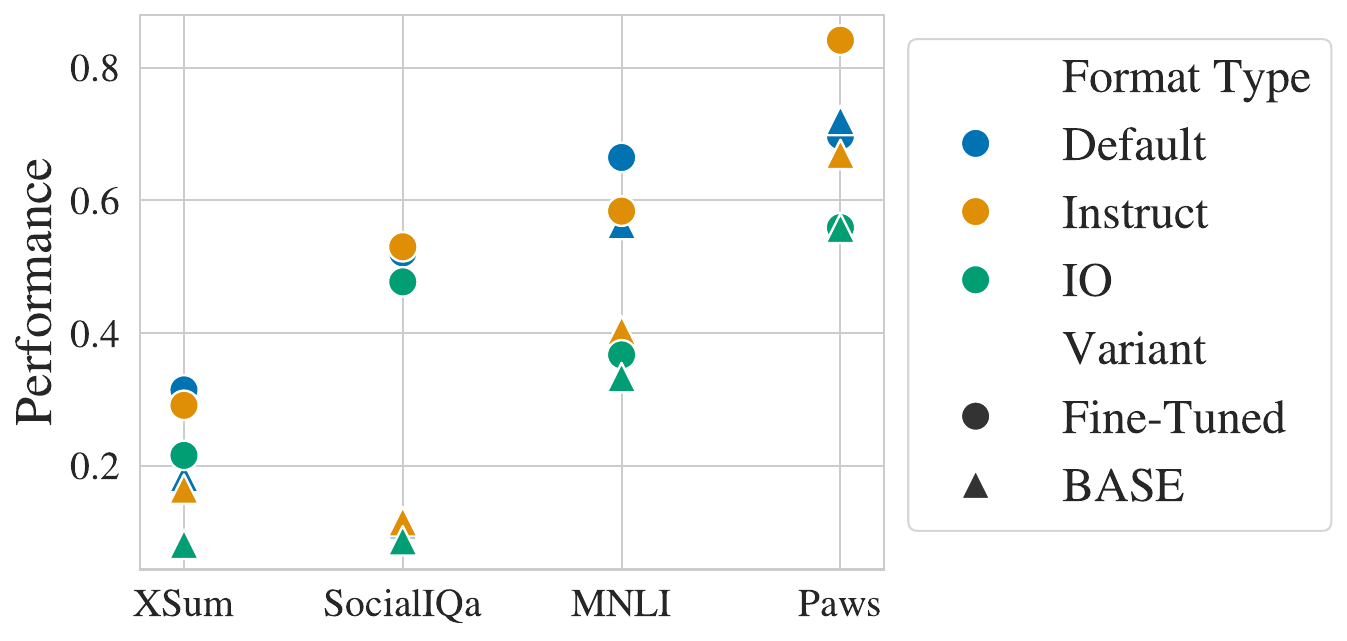}
  \caption{LLAMA3-8B performance with different task format. \texttt{Instruct} and \texttt{Default} always lead to highest evaluation results.}
  \label{fig:finding:task-format-llama}
  \vspace{-12pt}
\end{figure}

\textbf{Datasets.}
The datasets are summarized in Table ~\ref{tab:tasks} and data licenses are in 
Table~\ref{app:tab:artifact}.
We evaluate the model with an in-domain test set and one or two out-of-domain test sets for each of the supervised fine-tuning tasks.
We conduct experiments on the tasks of \textit{summary generation} \citep{narayan-etal-2018-dont, hasan-etal-2021-xl, hermann2015teaching}, \textit{question generation} \citep{sap-etal-2019-social, xiong-etal-2019-tweetqa, welbl2017crowdsourcing}, \textit{natural language inference} \citep{williams-etal-2018-broad, wang-etal-2018-glue, dagan2006pascal, bar2006second, giampiccolo2007third, bentivogli2009fifth}, and \textit{paraphrase detection} \citep{zhang-etal-2019-paws, wang-etal-2018-glue, agirre2007semantic}.
We subsample 6,000 training instances for each set to ensure a fair comparison.

In instruction fine-tuning, we base our downstream evaluation settings on \citet{groeneveld2024olmo}, as OLMo is found to have stable performance on these datasets.
The instruction-tuned models are evaluated on ARC (both \texttt{arc easy} and \texttt{arc challenge}) \citep{clark2018think}, OpenbookQA \citep{mihaylov-etal-2018-suit}, Hellaswag \citep{zellers-etal-2019-hellaswag}, BoolQ \citep{clark-etal-2019-boolq}, and SciQ \citep{welbl2017crowdsourcing}.

\textbf{Metrics.}
We use accuracy \citep{scikit-learn} for classification tasks and \textsc{Rouge-L} \citep{lin-2004-rouge} for generation tasks.
The maximum amount of newly generated tokens is set to 5 for classification tasks and 60 for generation tasks.
Outputs are generated with greedy decoding.
For classification tasks, we experiment with both constrained decoding and logit-based predictions.
We find the best performance by selecting the label with the highest logit of its first subtoken (Appendix~\ref{app:pred_gen}).

\section{Supervised Fine-Tuning: What does the model learn and forget?}
\label{sec:finding:what}
We begin our analysis with the supervised fine-tuning process to understand the downstream results of the training process.
Specifically, we explore three dimensions: \textbf{task format, task transfer, and domain knowledge}. In each case, we fine-tune both final checkpoints and intermediate pre-training checkpoints to understand the relationship between pre-training and fine-tuning.

\subsection{Task Format}
\label{sec:finding:what:taskformat}
LLMs can be extremely sensitive to prompt perturbation in few-shot settings \citep{sclar2023quantifying, leidinger2023language, salinas2024butterfly, wahle-etal-2024-paraphrase}.
We hypothesize that fine-tuning fits the model to a specific task format, resulting in higher performance when the evaluation set matches this format. 
To test this hypothesis, we vary the task format to either match the training format, use a different format, or rely on instructions.

We carefully construct three different prompt formats for the following settings. 1) 
\texttt{Default} is the same format used for training, where we expect the model to benefit from learning the task format;
2) \texttt{IO} format, by contrast, reflects a common way of performing supervised fine-tuning by incorporating only unprocessed input and output;
3) \texttt{Instruct} uses a human-readable instruction template to format the input.
Table~\ref{tab:app:promptformat} in the Appendix shows an example of each format.
The performance of Llama3-8B with different task formats is shown in Figure~\ref{fig:finding:task-format-llama}.
Checkpoint performance on OLMo before and after fine-tuning is shown in Figure~\ref{fig:findings:task_format}.

Across both models, \texttt{IO} format leads to the least favorable performance, as the only task-specific information in this format is included in the evaluation shots.
Model reports similar performance when evaluated with the \texttt{default} and \texttt{instruct} format, aligning with the findings in \citet{hewitt2024instruction} that the models retain their instruction-following ability after fine-tuning without instructions. 
However, in the early pre-training steps, aligning the task format with fine-tuning data plays a crucial role (Figure~\ref{fig:findings:task_format}), suggesting that the instruction-following ability has not yet been developed.
In this view, \textbf{fine-tuning teaches the model how to format a response for the task, while further pretraining enhances the instruction-following ability}. In other words, the instruction provides a directed prior for the model to behave in a certain way.

\begin{figure}[t!]
    \begin{subfigure}[b]{0.5\textwidth}
    \centering
    \includegraphics[width=0.7\columnwidth]{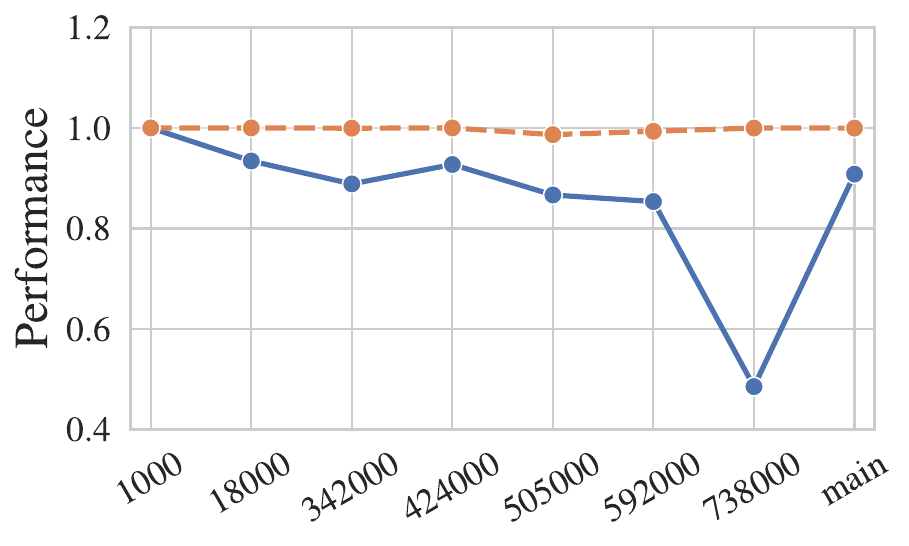}
        \caption{Paws $\rightarrow$ QQP}
        \label{fig:ood:detrimental}
    \end{subfigure}%
    \\
    \begin{subfigure}[b]{0.5\textwidth}
        \centering
    \includegraphics[width=0.7\columnwidth]{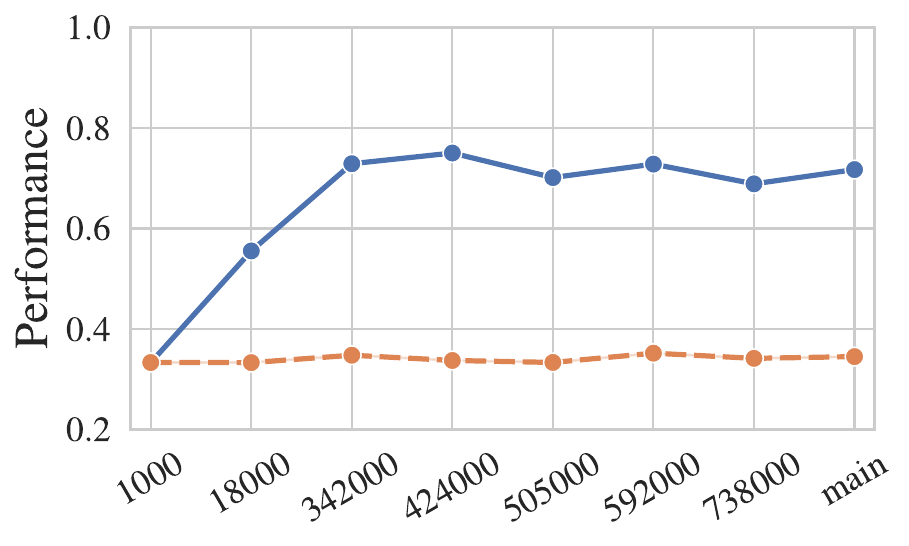}
        \caption{MNLI $\rightarrow$ GPT3NLI}
        \label{fig:ood:beneficial}
    \end{subfigure}
    \caption{Example of out-of-domain performance for fine-tuned models. The \textcolor{snsblue}{\textbf{solid blue}} line represents the fine-tuned checkpoint evaluated on an out-of-domain dataset, and the \textcolor{snsorange}{\textbf{dashed orange}} line represents the base checkpoint where the model is not fine-tuned. Figure~\ref{fig:ood:detrimental} shows an example of fine-tuning hurting OOD performance, while Figure~\ref{fig:ood:beneficial} shows an example of fine-tuning boosting OOD performance as pre-traininng proceeds. }
    \label{fig:findings:ood}
\end{figure}

\begin{figure}[t]
\centering
  \includegraphics[width=0.65\columnwidth]{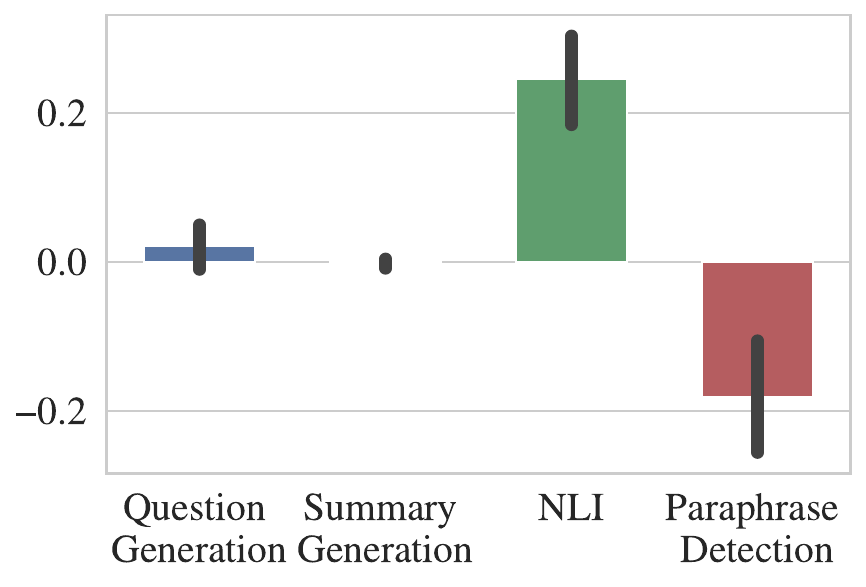}
  \caption{Ratio of out-of-domain performance change for each task, averaged across checkpoints.}
  \label{fig:finding:ood-by-data}
  \vspace{-12pt}
\end{figure}

\subsection{Domain Knowledge}
We next explore how the domain-generalization ability is affected by fine-tuning by inspecting whether the model forgets the domain knowledge after fine-tuning on a different domain.
An example of OOD model performance is shown in Figure~\ref{fig:findings:ood}, and the mean ratio of change by datasets is presented in Figure~\ref{fig:finding:ood-by-data} and Figure~\ref{fig:finding:ood-by-data-llama}.

The models do not benefit equally from the in-domain fine-tuning: Llama shows subtle benefits on question generation tasks, while not benefiting at all on the other tasks (Figure~\ref{fig:finding:ood-by-data-llama}). 
Across OLMo training history (Figure~\ref{fig:finding:ood-by-data}), NLI datasets experience a boost when fine-tuning on MNLI, while fine-tuning on Paws is detrimental to other paraphrase detection datasets.
This suggests that both forgetting and learning are happening in fine-tuning: the model learns to perform the task with in-domain knowledge, but it may, in turn, forget information more distant from what is learned in fine-tuning.
Furthermore, \textbf{under the same task, the amount of general-purpose pre-training may not affect the model's reaction to out-of-domain knowledge}.
Questions remain, however, about whether domain-specific continual pre-training or continual pretraining on similarly distributed data would bring different conclusions, which requires further study of pre-training dynamics.

\begin{figure*}[t!]
    \centering
    \begin{subfigure}[b]{0.47\textwidth}
    \includegraphics[width=\the\columnwidth]{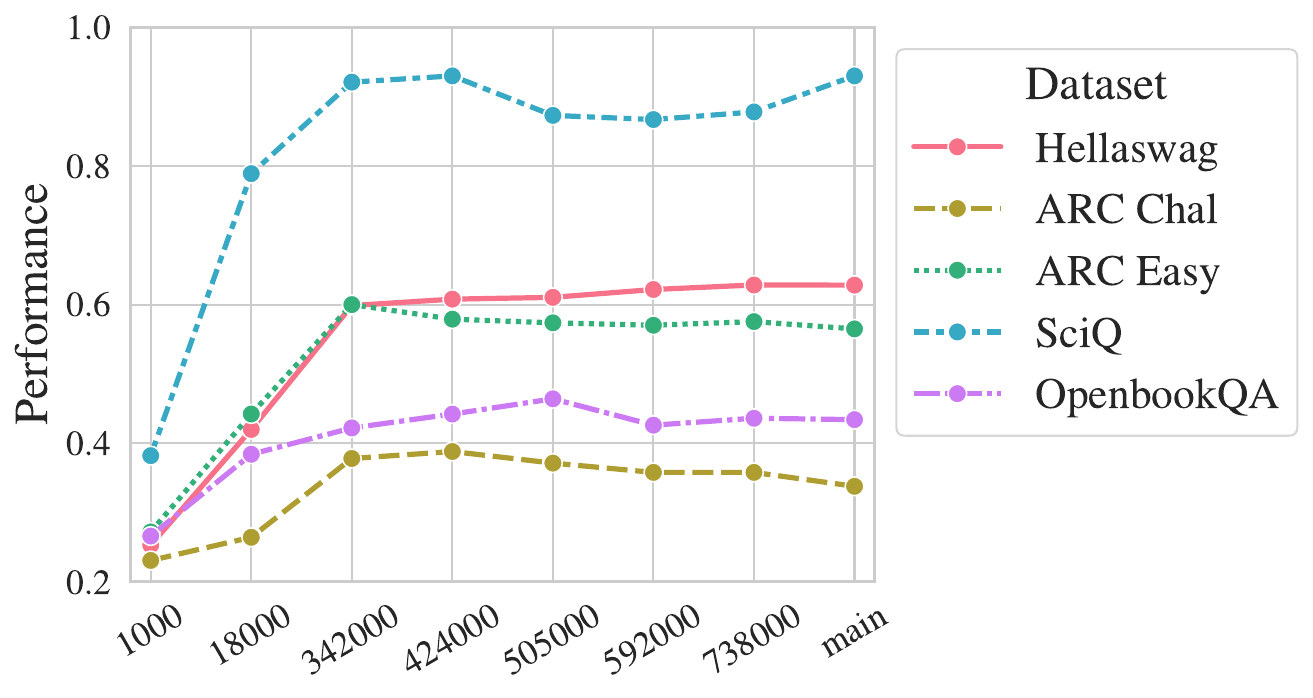}
        \caption{Learned during Pre-training.}
        \label{fig:base-eval-a}
    \end{subfigure}%
    ~ 
    \begin{subfigure}[b]{0.45\textwidth}
        \centering
    \includegraphics[width=\the\columnwidth]{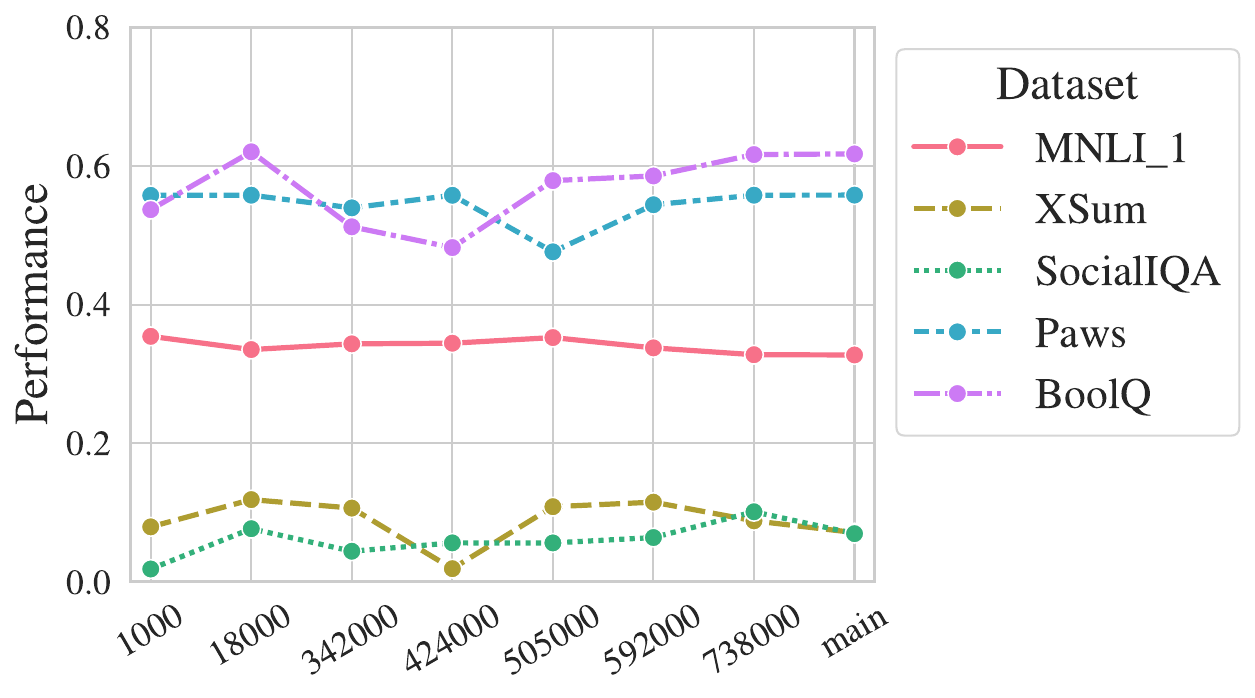}
        \caption{Never learned during pre-training.}
        \label{fig:base-eval-b}
    \end{subfigure}
    \caption{Few-shot performance on different pre-training steps.}
    \label{fig:base-eval}
    \vspace{-10pt}
\end{figure*}

\subsection{Task Transfer}
Model forgetting occurs when model training on new tasks improves those tasks at the expense of previously trained tasks \citep{luo2023empirical, mehta2023empirical, li2024examining}. 
To understand whether the model will forget a previously known task solution when fine-tuned on a different one, we evaluate model forgetfulness by examining whether the model does worse on some tasks after fine-tuning for other tasks. 
Specifically, we divide our tasks into two types: classification and generation. 

We notate the training datasets as $D_{T}$ and the evaluation datasets as $D_{E}$. 
We represent the performance of a pre-trained model (BASE) on checkpoint $i$ as ${\text{Perf}}_{BASE}^i(d)$ for an evaluation dataset $d\in D_{E}$, and the performance of the i-th checkpoint fine-tuned on dataset $t \in D_{T}$ be $\text{Perf}_{t}^{i}(d)$. To normalize the effect caused by uneven performance across different datasets, we compute the mean ratio of change (MRC) in performance for each checkpoint as follows.
\begin{equation}
\centering
    \resizebox{205pt}{!}{$\text{MRC} = \frac{1}{|D_{E}\setminus \{t\}|}\sum \limits_{\forall d \in D_{E}, d \neq t}{\frac{\text{Perf}_{t}^{i}(d) - {\text{Perf}}_{BASE}^i(d)}{\text{Perf}_{BASE}^i(d)}}$} 
\end{equation}

Models fine-tuned on classification tasks and evaluated on generation tasks decrease on average 61.4\% compared to models that are never fine-tuned.
In contrast, models fine-tuned on generation tasks can still perform the same as the BASE model on classification tasks, with a 0.3\% MRC, which is not statistically significantly different from a 0\% change.
Our findings on all pre-training checkpoints align with the findings of \citet{yang2024unveiling} on the final checkpoint of LLAMA-7B and our experiments on the final checkpoint of Llama3-8B (Appendix~\ref{app:sec:llama}).

Regardless of the pre-training stage, \textbf{a model maintains classification abilities when trained for generation but loses generation abilities when trained for classification.} 
This is not surprising given that classification tasks can be seen as a subset of generation, while the reverse is not true. 
The model follows a simplicity bias \citep{shah2020pitfalls} and thus is more likely to memorize simple classification tasks than generation tasks with an exponentially larger search space.
Additionally, since we evaluate the classification tasks based on the output logits and the base model performs randomly on the classification tasks, it is much easier for the models to maintain the same performance as the BASE models. 
Regardless of the stage of pre-training, fine-tuning can cause a model to lose abilities when the desired fine-tuning behavior does not support those abilities.

Across these three experimental settings, we find that fine-tuning teaches a model how to perform a task without hurting the model's instruction-following ability, but can sacrifice generalization across domains and tasks.

\section{How does the model change across pre-training?}

\label{sec:finding:base-eval}
Section~\ref{sec:finding:what:taskformat} reveals that the effect brought by fine-tuning could be different depending on the amount of pre-training, but how exactly does pre-training affect downstream fine-tuning results?
We begin by considering how additional pre-training changes the {\bf \texttt{BASE}} model. Typically, researchers track the value of the training or held-out loss during training. 
However, performance improvements on downstream tasks do not always follow the same trend with the loss curves \citep{groeneveld2024olmo}.

Instead, we evaluate the pre-trained checkpoints with few-shot examples, as models without alignment tend to do poorly in a zero-shot context.
Four shots are randomly sampled from the datasets, which are selected based on the highest performance shot amount reported in \citet{yang2024unveiling}. 
The model's performance at each pre-training step is reported in Figure~\ref{fig:base-eval}.

Broadly speaking, our results suggest that all datasets fall into one of two groups. 
For the first group of datasets (Figure~\ref{fig:base-eval-a}), although the model shows clear improvement during the early stages of pre-training, performance levels off fairly early on and remains consistent. The dramatic improvements in the early stages of pre-training may result from larger steps in early optimization.
We find improvements stop increasing past step 342,000.
The second group (Figure~\ref{fig:base-eval-a}) shows tasks that are never learned during pre-training. 
Performance remains constant throughout the whole pre-training process even when we vary shot sizes. 
These datasets include MNLI, XSum, and BoolQ.
A natural hypothesis for this finding is potential data contamination in the pre-training data.
However, the evaluation datasets are selected based on the popularity of the task and the content of pre-training data. 
All datasets that experience improvement do not exist in the model's pre-training data \citep{soldaini2024dolma}, while the more likely leaked datasets (MNLI, XSUM) never gain an improvement during the pre-training process.

Overall, these results reveal an interesting dichotomy. 
\textbf{Some tasks can be learned during pre-training, while others cannot}. 
Next, we explore what exactly the model is learning regarding this second group of datasets during pre-training by exploring the fine-tuned models.

\begin{figure}[t!]
    \begin{subfigure}[b]{0.5\textwidth}
    \centering
    \includegraphics[width=0.8\columnwidth]{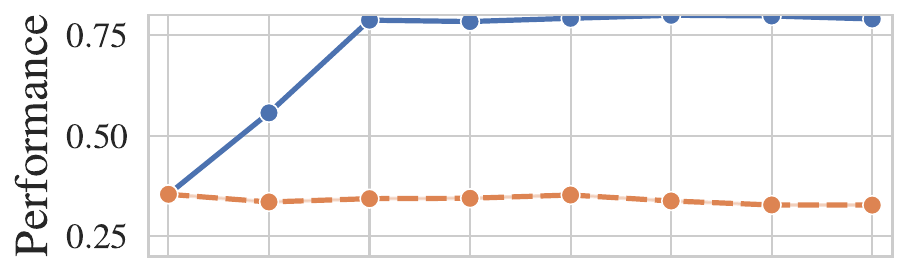}
        \caption{MNLI Matched}
        \label{fig:subfig:improve}
    \end{subfigure}%
    \\
    \begin{subfigure}[b]{0.5\textwidth}
        \centering
    \includegraphics[width=0.8\columnwidth]{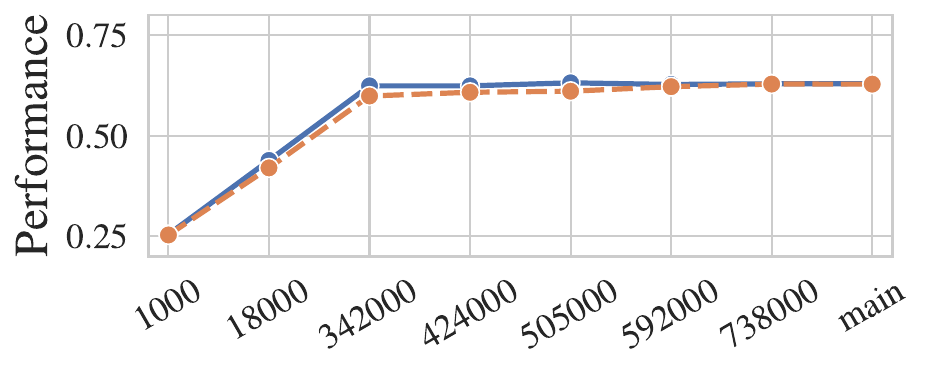}
        \caption{Hellaswag}
        \label{fig:subfig:notimprove}
    \end{subfigure}
    \caption{Example of few-shot performance on different pre-training steps of the models that benefited (\ref{fig:subfig:improve}) and did not benefit from fine-tuning (\ref{fig:subfig:notimprove}). The \textcolor{snsblue}{\textbf{solid blue}} line represents the fine-tuned checkpoint, and the \textcolor{snsorange}{\textbf{dashed orange}} line represents the base checkpoint. The results of all datasets can be found in Figure~\ref{fig:sft-ckpt-perf} and Figure~\ref{fig:it-ckpt-perf}.}
    \label{fig:improve-and-notimprove-example}
\end{figure}

\begin{figure}[t]
    \centering
  \includegraphics[width=0.84\columnwidth]{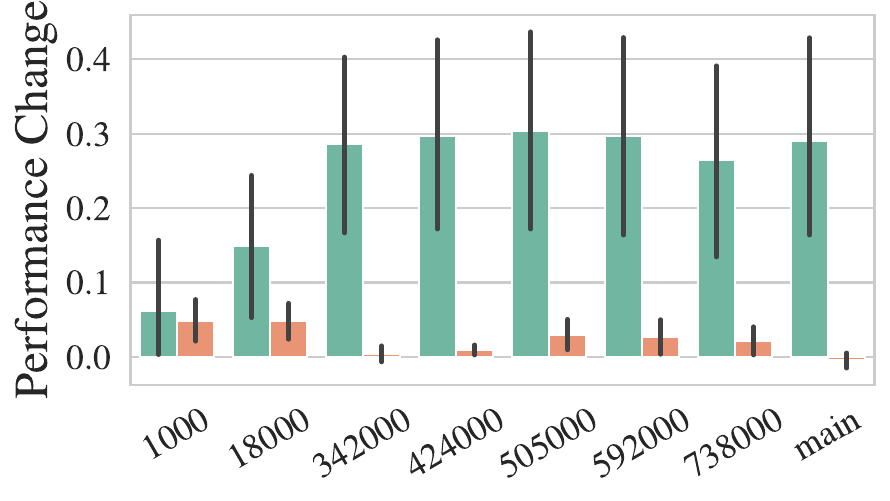}
  \caption{Amount of performance increase brought by fine-tuning between tasks that model can solve in pre-training (\textcolor{snslightorange}{\textbf{mandarin orange}}) and tasks that the model could not solve until fine-tuning (\textcolor{snsgreen}{\textbf{sage green}}). The exact number of mean increase is shown in Appendix~\ref{sec:app:performance-numbers}.}
  \label{fig:finding:ptftcompare}
\end{figure}

\section{Does more pre-training yield better fine-tuning results?}
\label{sec:finding:PTFT}
\citet{groeneveld2024olmo} compares OLMo's performance on several tasks before and after fine-tuning the final checkpoint and finds that fine-tuning enables the model to do well on tasks for which the unaligned model does poorly. We observe (\sect{sec:finding:base-eval}) that while some datasets improved 
during pre-training, there is a group of datasets for which a pre-trained model does poorly.
Does the model learn useful information for these tasks but cannot express it without fine-tuning? In this section, we further explore this dataset dichotomy by examining fine-tuned checkpoints for each of the datasets. 

Our results appear in Figure~\ref{fig:improve-and-notimprove-example} and Figure~\ref{fig:finding:ptftcompare}. First, we consider those datasets where the pre-trained models do well (Figure~\ref{fig:base-eval-a}). 
These datasets do not improve with fine-tuning, suggesting that whatever is learned during fine-tuning, which we discuss below, the model already gains the knowledge during pre-training. This effect is observed at all checkpoints; fine-tuning simply does not help.

However, a different story is observed for datasets that are not learned during pre-training. 
For these, fine-tuning yields significant improvements at every model checkpoint, with Figure~\ref{fig:finding:ptftcompare} showing the magnitude of improvement on these datasets compared to no improvement to the datasets already learned during pre-training.
Moreover, earlier checkpoints obtain more substantial gains from fine-tuning than later checkpoints.
The benefit of fine-tuning continues to increase until a certain threshold in pre-training steps is reached (approximately 424,000).
Figure~\ref{fig:improve-and-notimprove-example} shows representative plots comparing the performance of a pre-trained versus fine-tuned model at different checkpoints for two datasets (full list in Appendix~\ref{sec:app:per-ds-ckpt-figures}). 
For Hellaswag (learned during pre-training), fine-tuning does not benefit the model, even during early checkpoints when the model performs poorly on the task. 
Nevertheless, for MNLI (not learned during pre-training), fine-tuning dramatically improves the model. 
Interestingly, later checkpoints achieve better results after fine-tuning, even when the performance of the pre-trained model is unchanged. 
This suggests that \textbf{the model is, in fact, improving during pre-training, but it cannot express that improvement without fine-tuning}.

Our findings suggest that early stopping in pre-training will not be detrimental to downstream fine-tuning performance. When the budget is limited, the benefits of fine-tuning an LLM could exceed the benefits of continued pretraining, which sheds light on the potential of a cost-effective training paradigm with less pre-training.
However, directly identifying such stopping criteria without fine-tuning intermediate checkpoints is challenging.
We only empirically observed that the point where more pre-training lead to diminishing return on downstream fine-tuning results approximately align with the turning point of few-shot performance in Section~\ref{sec:finding:base-eval}.
Without such a hypothesis, the improvement trend is invisible before fine-tuning the checkpoints.
Overall, when resource-intensive pre-trained LLMs are not available, fine-tuning models on checkpoints with less pre-training may be a reasonable practical choice for obtaining a high-quality model.

\section{Discussion}
Our study fine-tunes model pre-training checkpoints to understand the dynamics of pre-training and fine-tuning on model performance. 

{\em Fine-tuning teaches additional task format but leads to forgetting unused abilities.}
Our results show that fine-tuning guides the model to understand the format and complete a given task. 
As this information diminishes, the model's overall ability improves. 
Additionally, more pre-training will lead to a model that reacts better to instruction-style prompts, and the ability to interpret such instruction will not be lost when the model is fine-tuned in a different format.
However, fine-tuning comes at the expense of other model abilities, such as the capability of solving tasks or domains that are unrelated or weakly related to the fine-tuning task. 
This insight can be helpful in our understanding of the multitask abilities of LLMs, where certain tasks can introduce conflicts during multi-task training \citep{mueller-etal-2022-text}.

{\em Some datasets can be learned without fine-tuning.}
We discover a dichotomy between datasets. 
Some are learned during model pre-training, while others show no improvements during pre-training. 
Furthermore, the datasets learned during pre-training do not benefit from fine-tuning. 
This observation, combined with our study about what is learned during fine-tuning (\sect{sec:finding:what}) suggests that some tasks are presented in a manner that aligns with what the model sees during pre-training, and thus fine-tuning provides no additional information. 
It may be possible to modify tasks to better align with pre-training and thus make them learnable.

{\em Pre-training can improve models in unseen ways.}
Some datasets are not learned during pre-training but benefit significantly from fine-tuning (\sect{sec:finding:base-eval}). 
However, these datasets still benefit from additional pre-training, even though those benefits are not revealed without fine-tuning (\sect{sec:finding:PTFT}).
The model learns important information to solve the task, even though it cannot express that information without fine-tuning.
We empirically observe that the point where more pre-training lead to diminishing return on downstream fine-tuning results approximately align with the turning point of few-shot performance in Section~\ref{sec:finding:base-eval}.
Future work may identify ways to verify the turning point and detect these improvements during pre-training, which can better guide pre-training choices to produce models that perform better post-fine-tuning. 
Perhaps there is a way in which information about these tasks can be included in pre-training, allowing the model to better utilize the massive amount of pre-training data.
For example, early stopping during pre-training could lead to better utilization of limited training resources if we know when to stop.

\section{Related Work}
\label{sec:relatedwork}
Recent studies identify \textit{phase transition} of model training \citep{olsson2022context, wei2022emergent}, where new capabilities or behaviors suddenly emerge when certain thresholds of model complexity are reached.
The aspects of complexity often include model size, amount of training by FLOPs or tokens, and model architecture.
Several prior works studied the training dynamics of language models by analyzing the internals of train-from-scratch models \citep{tirumala2022memorization, chen2023sudden, tian2023scan, chen2024training, chang2024characterizing}.
The results of these works suggest that the behaviors that are often overlooked after training could be valuable signals for model analysis.
In addition to train-from-scratch models, \citet{ren2024learning} studied the fine-tuning dynamics of language models.
This work focuses on \textit{the effect of pre-training dynamics on downstream fine-tuning results} by fine-tuning intermediate pre-training checkpoints on various tasks.
Due to the scarcity of publically accessible intermediate pre-training checkpoints, the effect of fine-tuning at different pre-training stages is largely unexplored.
Concurrent work \citep{snell2024predicting} also fine-tunes intermediate pre-training checkpoints and finds that supervised fine-tuning results can be used as a signal to predict when emergence occurs, while our findings point out a dichotomy of model behavior on different datasets, with the potential for data-efficient and budget-friendly training by understanding the stages of model training.

\section{Conclusion}
We explore the relationship between fine-tuning and pre-training LLMs through fine-tuning multiple pre-training checkpoints of large language models.
Our results on 18 datasets and two models provide insights into LLM training.
We identify the aspects that LLM learns and forgets during supervised fine-tuning;
By analyzing pre-training history, we find that pre-training improves the model in a latent way that is only observable after fine-tuning.
The model may excel at some tasks without fine-tuning.
However, the model can rapidly learn datasets that it does not demonstrate capabilities during pre-training with a small amount of supervision.
Overall, our study highlights the value of analyzing language model training dynamics.
We encourage model developers to release pre-training checkpoints to facilitate research on LLM training.

\section*{Limitations}
\label{sec:limitations}
While our insights suggest directions for future work, we note important limitations inherent in our experiments. 
We discuss the weaknesses and limitations in the following section.

\paragraph{Computing Resource. }
Due to computational constraints, we can only conduct checkpointing experiments on a 1B model.
We supply the final checkpoint of an 8B model to verify the findings that are shared across checkpoints.
The amount of GPU hours spent for each experiment in this study is listed in Table~\ref{app:tab:GPU-hours}.

\paragraph{Model Size and Variant. }
For the analysis with intermediate checkpoints, this study considered a single, relatively small LLM, which may, therefore, conceal the emergent capability brought by larger models \citep{wei2022emergent}.
To combat this, we include the final checkpoint of an 8B model from a different model family.
Future work needs to confront these issues on larger models and more datasets.

\paragraph{Availbility of Pre-training Checkpoints. }
Although \citet{choshen2024hitchhiker} points out that the behavior of a model can often be predicted with a model with the same architecture but a different family.
This study would benefit significantly from including a broader spectrum of models, but the public pre-training checkpoint releases are limited.
We list open-source LLMs with intermediate checkpoint release in Appendix~\ref{sec:app:modelselection}.
After a series of preliminary experiments, we select available models' best-performing and robust families.

\paragraph{Analysis Protocol. }
\citet{wu2023reasoning} show that the evaluation result may be affected by samples that have been memorized by the model during training instead of revealing the reasoning capability.
The only analysis protocol used in this work is the downstream performance of a trained model.
More investigation should be done into model internals during pre-training dynamics and how they relate to the effects of fine-tuning.

\paragraph{Training Paradigm. }
Although multiple tuning strategies exist, to create a fair comparison environment where checkpoints receive the same amount of training, models are fine-tuned with a fixed amount of epochs in this work.
On different pre-training stages, the model may converge at a different speed.
Further study can be done to study the effect of pre-training on different fine-tuning methods or fine-tuning dynamics in different pre-training stages.
We only explored the scenario of full-parameter fine-tuning.
Whether parameter-efficient fine-tuning or human preference tuning will lead to a different conclusion also remains an open question.

\paragraph{Randomness. }
In this study, we only assess uncertainty with Bootstrap during evaluation.
However, uncertainty may emerge during training, which poses optimizer initialization and data ordering, the study of which requires an extensive amount of computing resources.

\section*{Acknowledgments}
The authors thank Saleh Soltan, Niyati Bafna, Fan Bai, Miriam Wanner, Xinbo Wu, and Carlos Aguirre for their helpful feedback.
This work was supported in part by a grant from the JHU + Amazon Initiative for Interactive AI.

\bibliography{custom,anthology}

\appendix
\section{Model and Data Selection}
\label{sec:app:modelselection}
\begin{table*}[ht]
\centering
\small
\begin{tabular}{@{}llllll@{}}
\toprule
 & \multicolumn{1}{c}{\textbf{Pythia}} & \multicolumn{1}{c}{\textbf{OpenLLAMA}} & \multicolumn{1}{c}{\textbf{K2 (LLM360)}} & \multicolumn{1}{c}{\textbf{Crystal (LLM360)}} & \multicolumn{1}{c}{\textbf{Baichuan2}} \\
  \midrule
\textbf{Citation} & \multicolumn{1}{l}{\citealp{biderman2023pythia}} & \multicolumn{1}{l}{\citealp{openlm2023openllama}} & \multicolumn{1}{l}{\citealp{k2model}} & \multicolumn{1}{l}{\citealp{tao2024crystal}} & \multicolumn{1}{l}{\citealp{yang2023baichuan}} \\ \\
\textbf{Size (Param)} & \begin{tabular}[c]{@{}l@{}}70M, 160M, \\ 410M, 1B, \\ 1.4B, 2.8B, \\ 6.9B, 12B\end{tabular} & 3B, 7B & 65B & 7B & 7B, 13B \\
\\
\textbf{Languages} & English & English & English & English & English \& Chinese \\
\\
\textbf{\begin{tabular}[c]{@{}l@{}}Pre-trained \\ Tokens\end{tabular}} & 300B & 1T & 1.4T & 1300B & 2.6T \\
\textbf{Note} & - & - & \begin{tabular}[c]{@{}l@{}}Multi-phase \\ pre-training\end{tabular} & \begin{tabular}[c]{@{}l@{}}Multi-phase \\ pre-training\end{tabular} & - \\ \\
\toprule
 & \multicolumn{1}{c}{\textbf{OLMO-2}} & \multicolumn{1}{c}{\textbf{OLMO}} & \multicolumn{1}{c}{\textbf{TinyLLaMA}} & \multicolumn{1}{c}{\textbf{RedPajama-INCITE}} & \multicolumn{1}{c}{\textbf{Bloom}} \\
 \midrule
\textbf{Citation} & \multicolumn{1}{l}{\citealp{olmo2}} & \multicolumn{1}{l}{\citealp{groeneveld2024olmo}} & \multicolumn{1}{l}{\citealp{zhang2024tinyllama}} & \multicolumn{1}{l}{\citealp{together2023redpajama}} & \multicolumn{1}{l}{\citealp{le2023bloom}} \\ \\
\textbf{Size (Param)} & 4T, 5T & 1B, 7B & 1B & 7B & 176B \\
\\
\textbf{Languages} & English & English & English & English & Multilingual \\ \\
\textbf{\begin{tabular}[c]{@{}l@{}}Pre-trained \\ Tokens\end{tabular}} & 7B, 13B & 3T, 2.5T & 3T & 1.2T & 366B \\
\textbf{Note} & \begin{tabular}[c]{@{}l@{}}Multi-phase \\ pre-training\end{tabular} & - & \begin{tabular}[c]{@{}l@{}}BOS Token leads \\ to \href{https://github.com/jzhang38/TinyLlama/issues/83}{training history} \\ inconsistency.\end{tabular} & \begin{tabular}[c]{@{}l@{}}Poor \\ fine-tunablity\end{tabular} & - \\
\bottomrule
\end{tabular}
\caption{Large language models with public release of intermediate pre-training checkpoints. All models are under Apache 2.0 license.}
\label{tab-app-checkpoints}
\end{table*}

Only a small subset of large language models publicly release their intermediate training checkpoints.
We list these models in Table~\ref{tab-app-checkpoints} and would like to call for model developers to release intermediate checkpoints in the future to aid the research of training dynamics.
To reduce the confounding factor of language and stages of training, we select the models that are dominantly trained in English and followed a single-staged training strategy.
Only the models pre-trained with significantly more tokens than the model parameters are considered to avoid the occurrence of double descent \citep{belkin2019reconciling, schaeffer2023double} in the middle of pre-trianing, which could lead to a broken scaling law \citep{caballero2022broken} that complicates the analysis.
Additionally, we restrict our selection to models pre-trained on over one trillion tokens, thereby ensuring a sufficiently extended training trajectory is represented.
We conduct initial experiments with OLMo and RedPajama-INCITE. 
We observe that the RedPajama-INCITE shows subtle improvement following instruction-tuning or fine-tuning, and its 7B variant shows lower performance compared to OLMo 1B.
Therefore, we select OLMo 1.0 1B as the backbone for analysis.

During this study, several recent initiatives released the intermediate checkpoints.
We also list these works in Table~\ref{tab-app-checkpoints}.

\section{Hyperparameter Tuning}
\label{sec:app:hyperparameter}
For both supervised fine-tuning and instruction tuning, we pre-set the effective batch size to 8, and tune the learning rate within \{$2\times 10^{-5}$, $2\times 10^{-6}$, $2\times 10^{-7}$\}.
OLMo-1B is fine-tuned for 3 epochs on the supervised fine-tuning tasks and 5 epochs on Tulu for instruction tuning.
Llama3-8B is fine-tuned for 2 epochs with a learning rate of $5\times 10^{-6}$, with learning rate selected from \{$5\times 10^{-5}$, $5\times 10^{-6}$, $5\times 10^{-7}$\}.
In both settings, we adopt an AdamW optimizer with a linear learning rate scheduler.
The optimizer is warmed up for the first $3\%$ of the training time.

\section{Prediction Generation Method}
\label{app:pred_gen}
\begin{table}[ht]
\small
\centering
\begin{tabular}{@{}llrrr@{}}
\toprule
\textbf{Dataset}       & \textbf{Model} & \multicolumn{1}{l}{\textbf{Free}} & \multicolumn{1}{l}{\textbf{Constrained}} & \multicolumn{1}{l}{\textbf{TokenProb}} \\ \midrule
\multirow{2}{*}{MNLI}  & Fine-tuned     & 0.786                                        & 0.791                                             & 0.792                                         \\
                       & BASE           & 0.0                                          & 0.0                                               & 0.327                                         \\ \arrayrulecolor{black!30}\midrule
\multirow{2}{*}{RTE}   & Fine-tuned     & 0.658                                        & 0.662                                             & 0.66                                          \\
                       & BASE           & 0.0                                          & 0.0                                               & 0.241                                         \\ \arrayrulecolor{black!30}\midrule
\multirow{2}{*}{Paws}  & Fine-tuned     & 0.871                                        & 0.878                                             & 0.878                                         \\
                       & BASE           & 0.0                                          & 0.0                                               & 0.558                                         \\ \arrayrulecolor{black!30}\midrule
\multirow{2}{*}{STS-B} & Fine-tuned     & 0.775                                        & 0.741                                             & 0.744                                         \\
                       & BASE           & 0.0                                          & 0.0                                               & 0.964                                         \\ \arrayrulecolor{black}\bottomrule
\end{tabular}
\caption{Performance of final checkpoint with different prediction generation method.}
\label{app:tab:pred_gen_method}
\end{table}

For classification tasks, we examine three different prediction generation methods: Free Generation (\texttt{Free}), Constrained Generation (\texttt{Constrained}), and Token Probability (\texttt{TokenProb}), the results are shown in Table~\ref{app:tab:pred_gen_method}.
In \texttt{Constrained}, we force the output to include at least one label in the acceptable label set.
In \texttt{TokenProb}, we compare the logits of acceptable labels and select the label with the highest score as the final output.
This ablation study is conducted only on the BASE and fine-tuned versions of the final checkpoint of the pre-trained model.
We find that, although prediction generation methods have less effect on the evaluation result of a fine-tuned model, BASE variants suffer much more from not knowing the desired output.
Therefore, we proceed with all the classification experiments with \texttt{TokenProb}.

\subsection{Label and Tokenizations}
Depending on the tokenizer variant, the label text may be tokenized differently, leading to evaluation unreliability.
For example, in paraphrase detection, the model could assign probability on both ``\texttt{yes}" and ``\texttt{ yes}" (the same label with a prefix space).
This behavior is reported and explored in various related work \citep{sun-etal-2023-tokenization, batsuren2024evaluating, singh2024tokenization}.
In this study, we leniently regard all individual tokens that contain the whole label or part of the label along with some special characters that do not affect the semantics as an acceptable target label.

\section{Task Format}
We adopt the task format from \citep{yang2024unveiling}, with an additional task format of input-output.
How each dataset is formated can be found in Table~\ref{tab:app:promptformat}.

\section{GPU Hours per-Experiment}
\label{sec:app:gpuhours}
We show a table of GPU hours spent for each experiment in Table~\ref{app:tab:GPU-hours}.
The total number of GPU hours spent on this project is approximately 1067 A100 hours. We lose track of the GPU hours spent on preliminary experiments, so a lower-bound estimation is reported.
\begin{table*}[ht]
\centering
\small
\begin{tabular}{@{}crrrr@{}}
\toprule
\multicolumn{5}{c}{\textbf{Prelinminary Experiments}} \\ \midrule
\multicolumn{3}{l}{\textbf{Description}} & \multicolumn{2}{l}{\textbf{GPU Hours}} \\ \arrayrulecolor{black!30}\midrule
\multicolumn{3}{l}{Instruction Tuning on LIMA, TULU, and NaturalInstructions} & \multicolumn{2}{r}{$\sim$300} \\
\multicolumn{3}{l}{Model Performance Verification, Dataset Selection} & \multicolumn{2}{r}{120} \\ \arrayrulecolor{black}\midrule
\multicolumn{5}{c}{\textbf{Instruction Tuning}} \\ \arrayrulecolor{black}\midrule
\multicolumn{3}{l}{Instruction Tuning} & \multicolumn{2}{r}{360} \\
\multicolumn{3}{l}{Evaluation} & \multicolumn{2}{r}{10} \\
\multicolumn{3}{l}{Total} & \multicolumn{2}{r}{\textbf{370}} \\ \arrayrulecolor{black}\midrule
\multicolumn{5}{c}{\textbf{Fine-Tuning}} \\ \arrayrulecolor{black!30}\midrule
\multicolumn{1}{l}{} & \multicolumn{1}{l}{\textbf{XSum}} & \multicolumn{1}{r}{\textbf{SocialIQa}} & \multicolumn{1}{l}{\textbf{MNLI}} & \multicolumn{1}{l}{\textbf{Paws}} \\ \arrayrulecolor{black}\midrule
\textbf{Training} & 12 & 6 & 4.6 & 5.3 \\
\textbf{Evaluation} & 8 & 5.3 & 3 & 2 \\
\textbf{OOD Evaluation} & 96 & 32 & 11 & 25.6 \\
\textbf{CrossTask Evauation} & 5.2 & 6.5 & 7.7 & 8.15 \\
\textbf{Task Format Evaluation} & 16 & 12.8 & 6 & 4 \\
\textbf{Total} & \multicolumn{4}{c}{137.2 + 62.6 + 32.3 + 45 = 277.1} \\ \arrayrulecolor{black}\bottomrule
\end{tabular}
\caption{GPU hours for each experiment. The total amount of GPU hours spent in this project is approximately 1067 A100 hours.}
\label{app:tab:GPU-hours}
\end{table*}

\section{Per-dataset Figures}
\label{sec:app:per-ds-ckpt-figures}
\begin{figure*}[t!]
    \centering
    \begin{subfigure}[b]{0.3\textwidth}
    \includegraphics[width=\the\columnwidth]{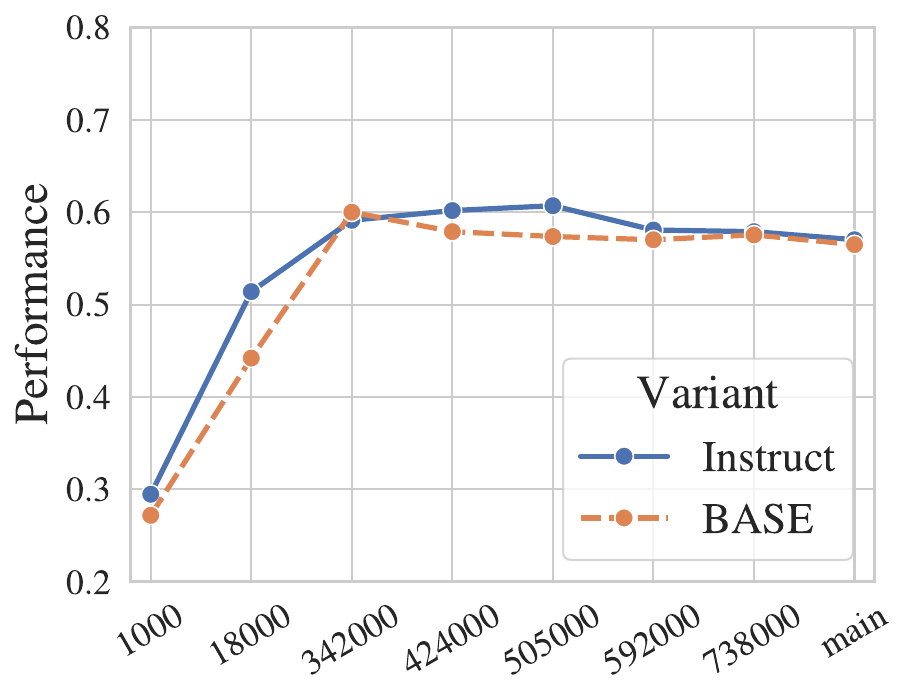}
        \caption{ARC Easy}
    \end{subfigure}%
    ~ 
    \begin{subfigure}[b]{0.3\textwidth}
    \includegraphics[width=\the\columnwidth]{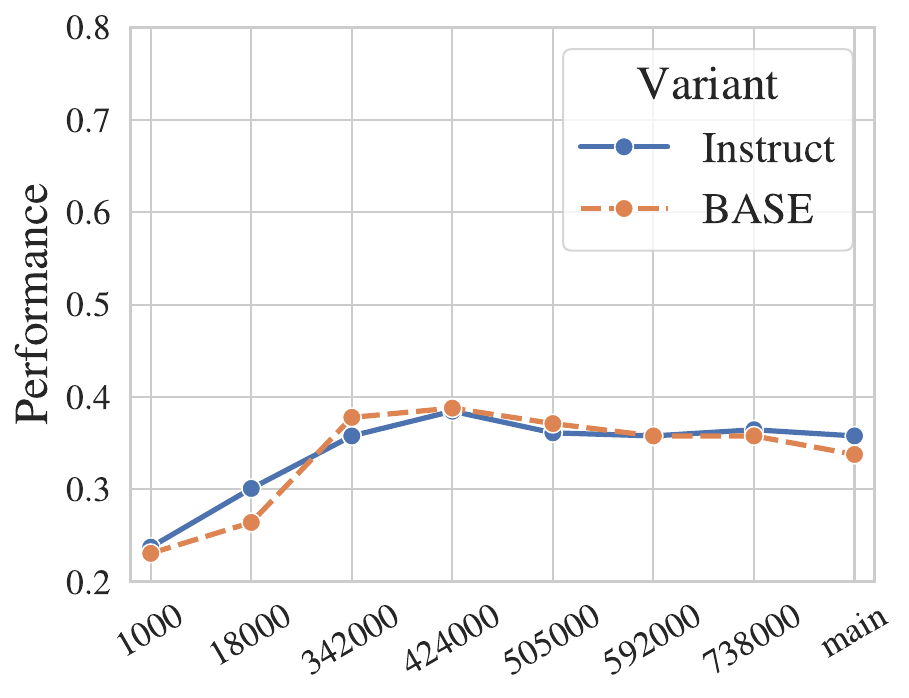}
        \caption{ARC Challenge}
    \end{subfigure}%
    ~ 
    \begin{subfigure}[b]{0.3\textwidth}
    \includegraphics[width=\the\columnwidth]{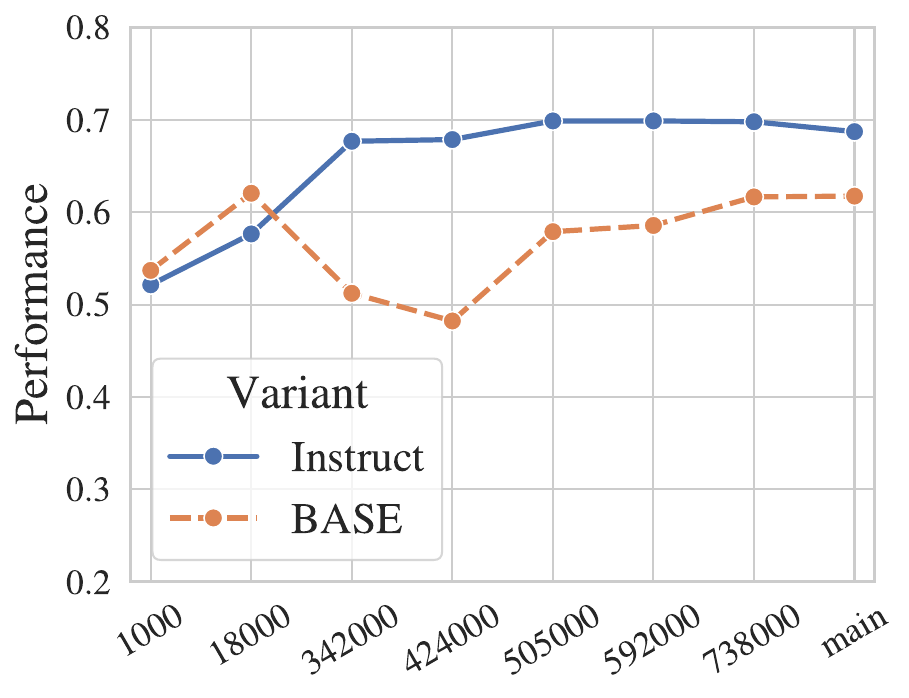}
        \caption{BoolQ}
    \end{subfigure}%
    \\
    \begin{subfigure}[b]{0.3\textwidth}
    \includegraphics[width=\the\columnwidth]{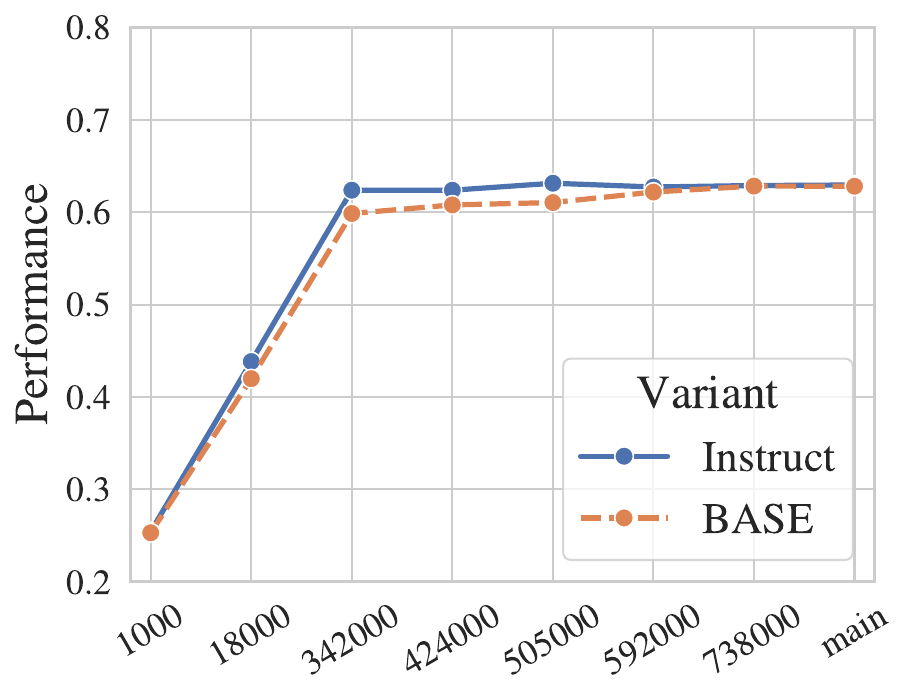}
        \caption{Hellaswag}
    \end{subfigure}%
    ~ 
    \begin{subfigure}[b]{0.3\textwidth}
    \includegraphics[width=\the\columnwidth]{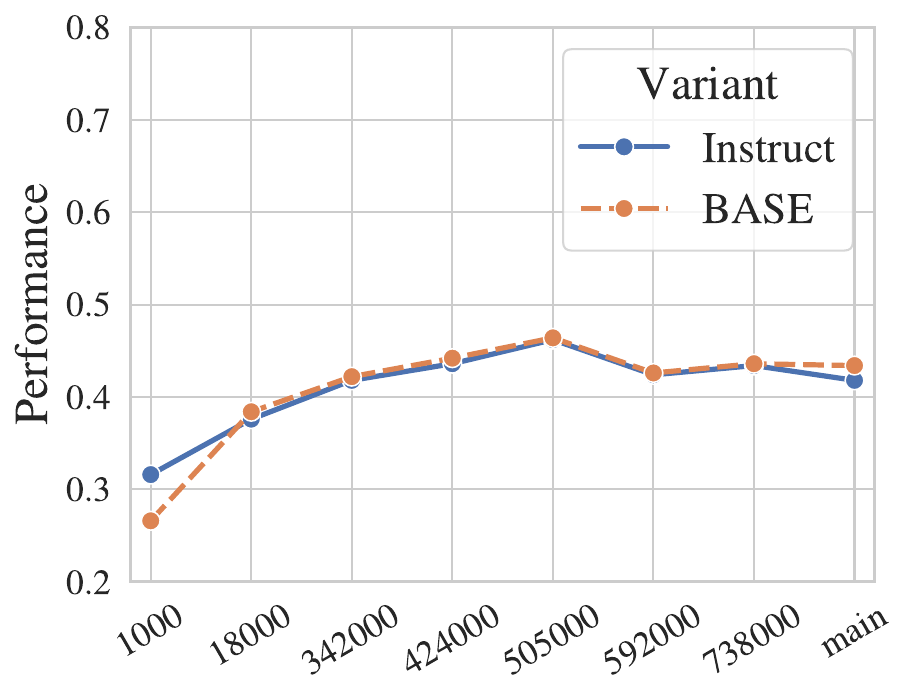}
        \caption{Openbook QA}
    \end{subfigure}%
    ~ 
    \begin{subfigure}[b]{0.3\textwidth}
    \includegraphics[width=\the\columnwidth]{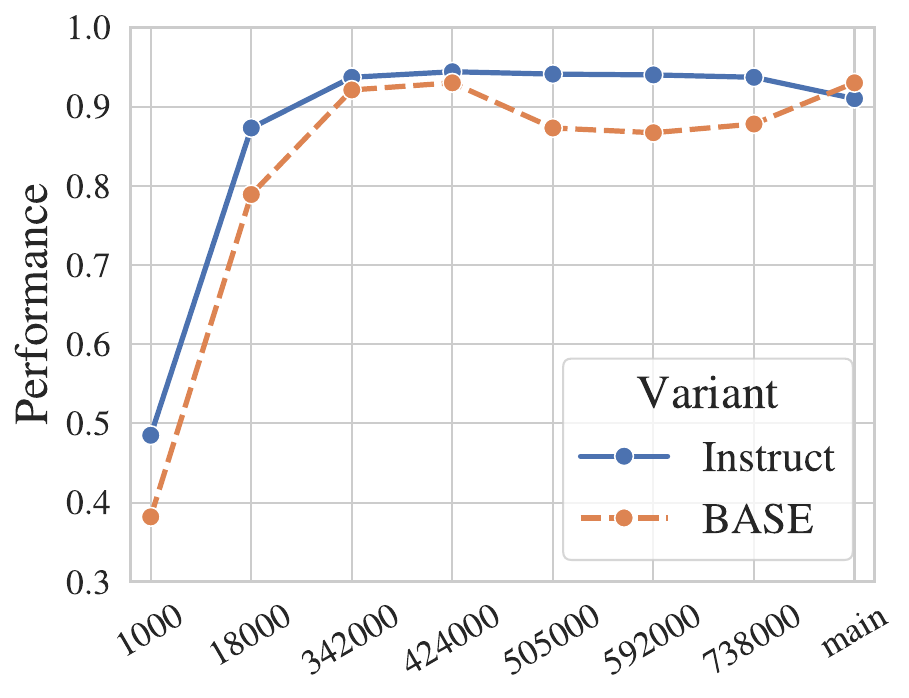}
        \caption{SciQ}
    \end{subfigure}%
    \caption{Model performance after instruction tuning on each pre-training step.}
    \label{fig:it-ckpt-perf}
\end{figure*}

\begin{figure*}[t!]
    \centering
    \begin{subfigure}[b]{0.3\textwidth}
    \includegraphics[width=\the\columnwidth]{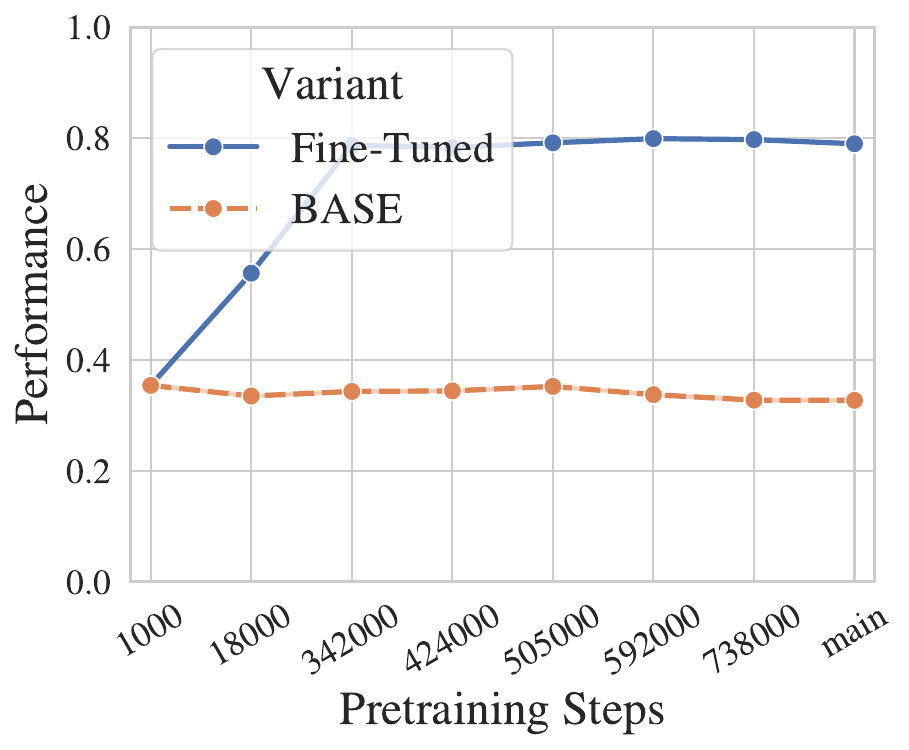}
        \caption{MNLI matched}
    \end{subfigure}%
    ~ 
    \begin{subfigure}[b]{0.3\textwidth}
    \includegraphics[width=\the\columnwidth]{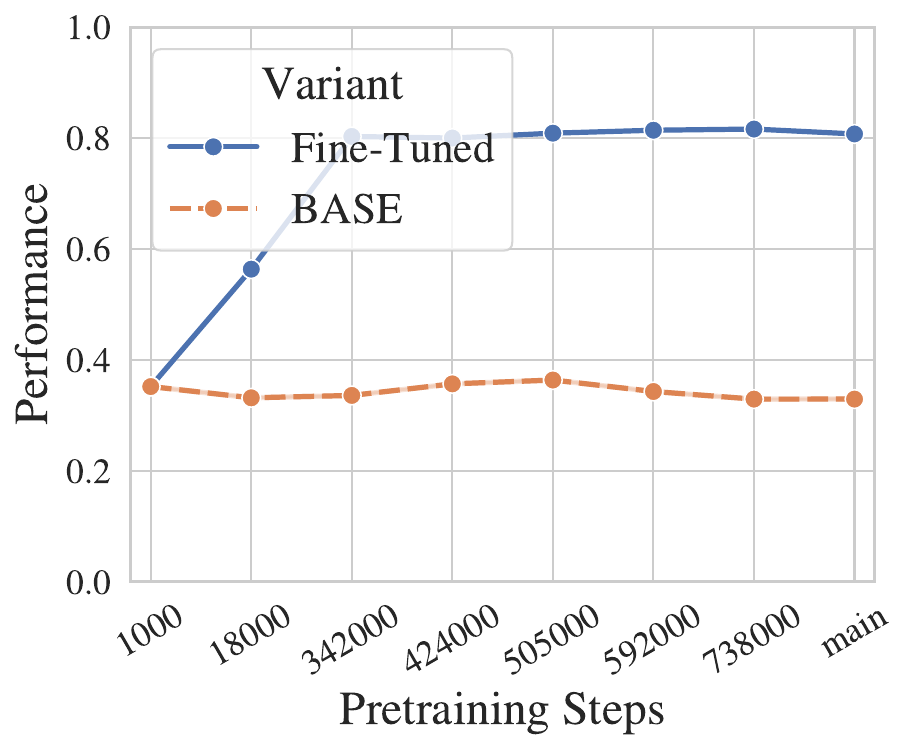}
        \caption{MNLI mismatched}
    \end{subfigure}%
    ~ 
    \begin{subfigure}[b]{0.3\textwidth}
    \includegraphics[width=\the\columnwidth]{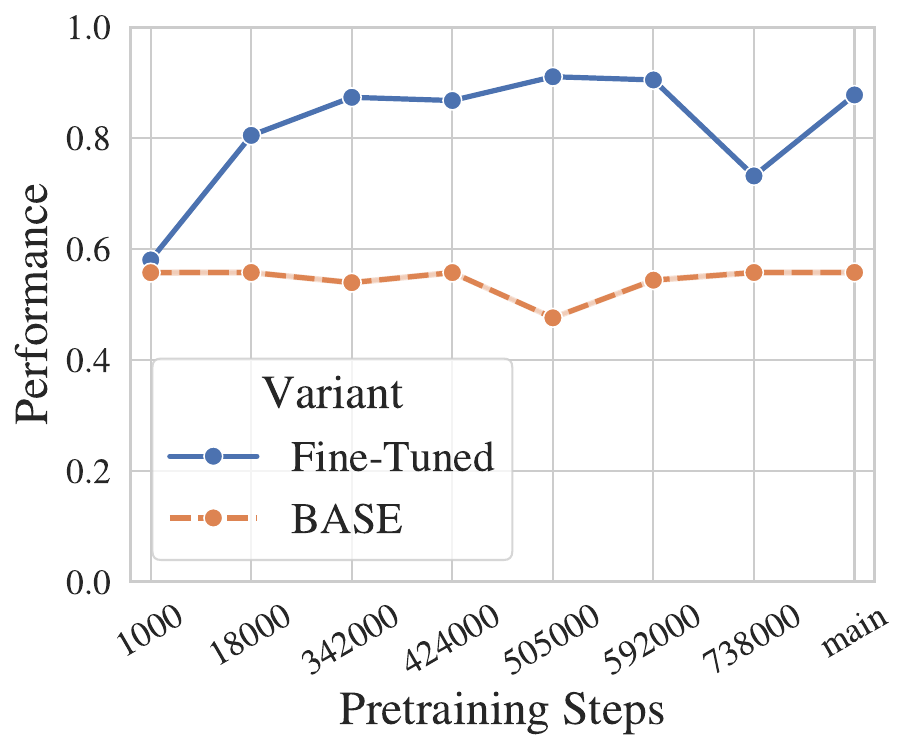}
        \caption{Paws}
    \end{subfigure}%
    \\
    \begin{subfigure}[b]{0.3\textwidth}
    \includegraphics[width=\the\columnwidth]{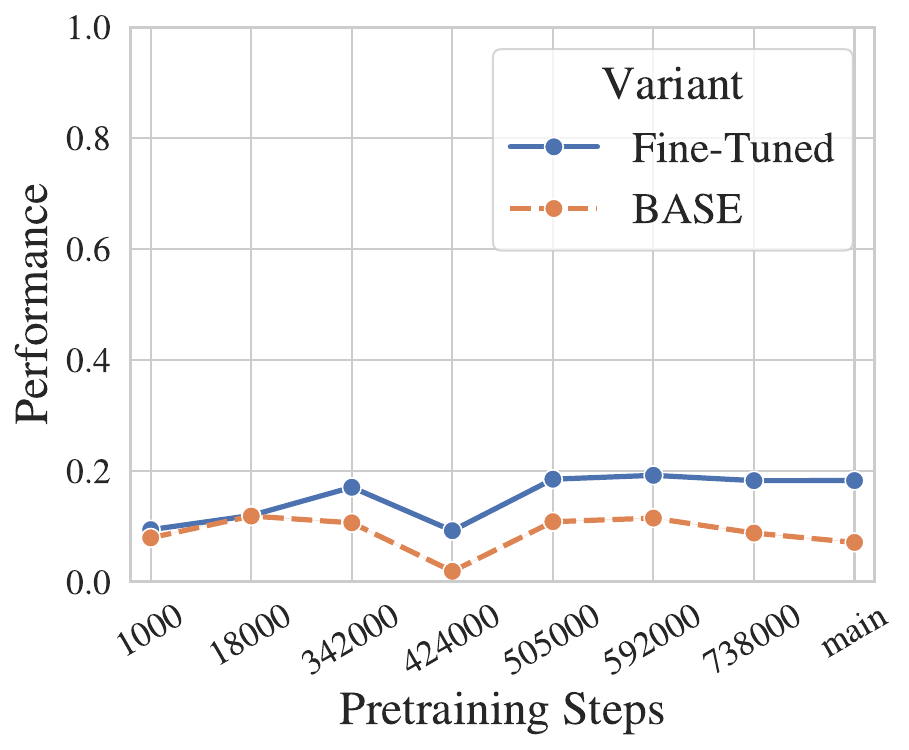}
        \caption{XSum}
    \end{subfigure}%
    ~ 
    \begin{subfigure}[b]{0.3\textwidth}
    \includegraphics[width=\the\columnwidth]{figures/fig_files/ft_ckpts/sft_evalxsum-trainxsum.pdf}
        \caption{XLSum}
    \end{subfigure}%
    ~ 
    \begin{subfigure}[b]{0.3\textwidth}
    \includegraphics[width=\the\columnwidth]{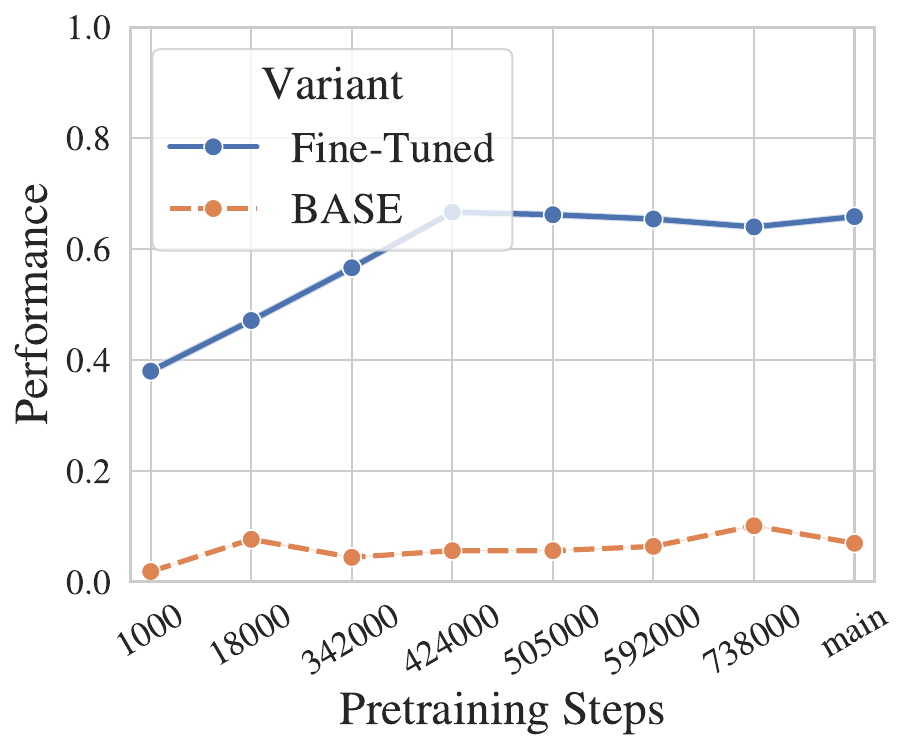}
        \caption{SocialIQA}
    \end{subfigure}%
    \caption{Model performance after supervised fine-tuning on each pre-training step.}
    \label{fig:sft-ckpt-perf}
\end{figure*}

We show the model performance on each dataset after supervised fine-tuning and instruction tuning correspondingly in Figure~\ref{fig:sft-ckpt-perf} and Figure~\ref{fig:it-ckpt-perf}.
The datasets that already show improvement during pre-training do not benefit from fine-tuning, while performance improves drastically on the datasets that the model has never learned during pre-training.

\begin{figure*}[t!]
    \centering
    \begin{subfigure}[b]{0.25\textwidth}
    \includegraphics[width=\the\columnwidth]{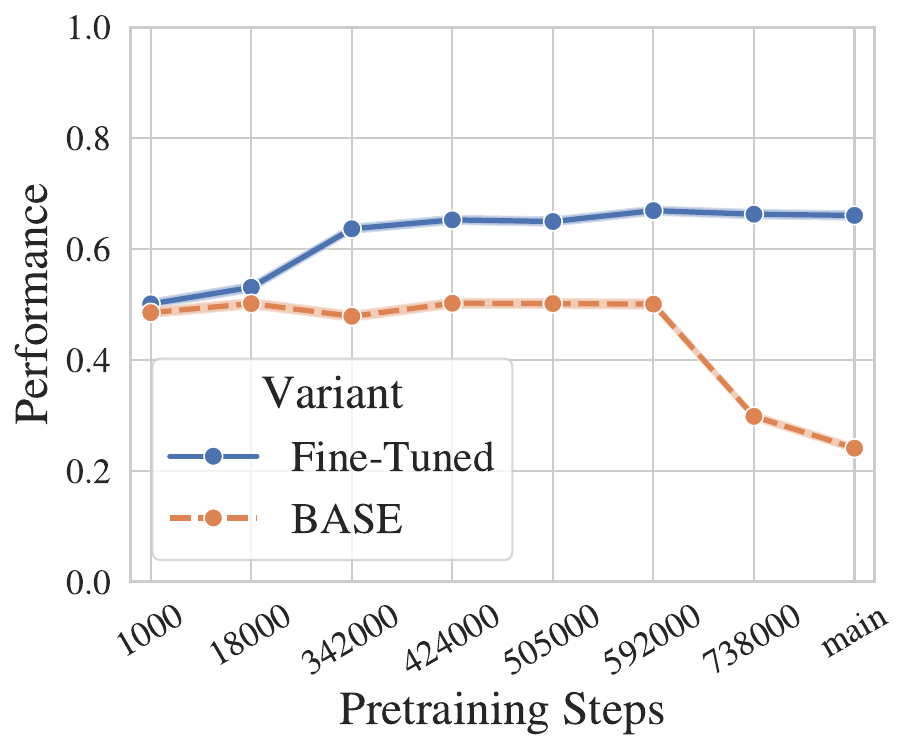}
        \caption{MNLI -> RTE}
    \end{subfigure}%
    ~ 
    \begin{subfigure}[b]{0.25\textwidth}
    \includegraphics[width=\the\columnwidth]{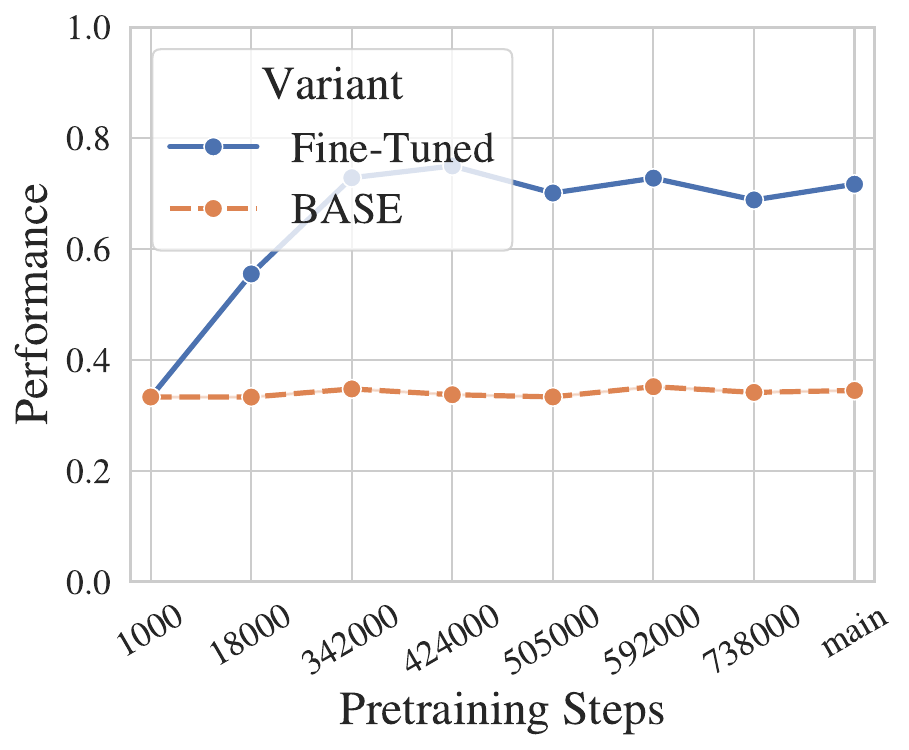}
        \caption{MNLI -> GPT3NLI}
    \end{subfigure}%
    ~ 
    \begin{subfigure}[b]{0.25\textwidth}
    \includegraphics[width=\the\columnwidth]{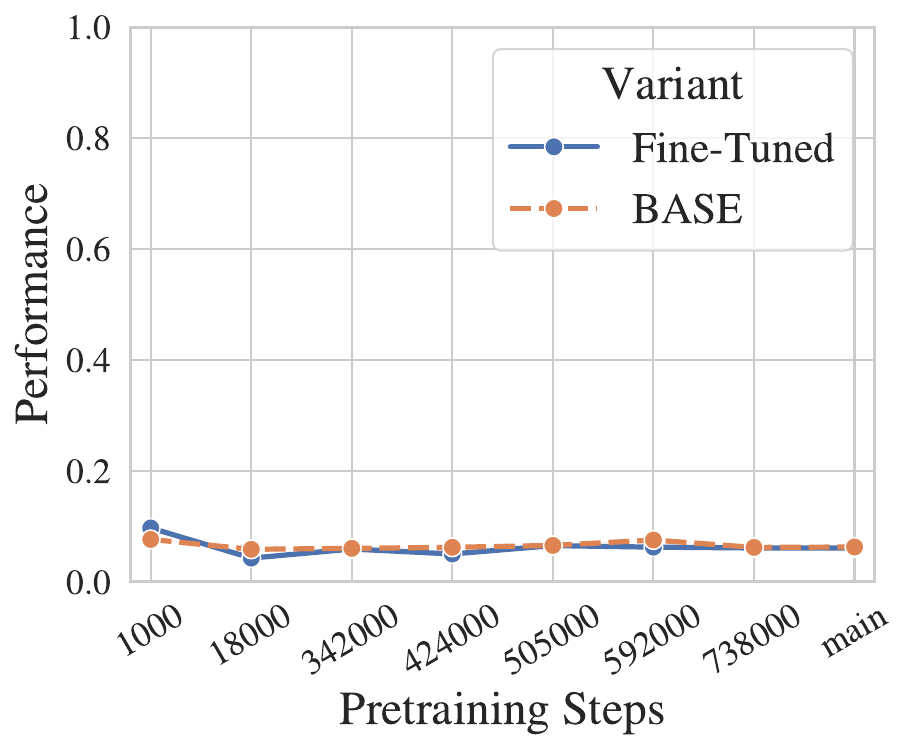}
        \caption{XSum -> CNN}
    \end{subfigure}%
    \\
    \begin{subfigure}[b]{0.25\textwidth}
    \includegraphics[width=\the\columnwidth]{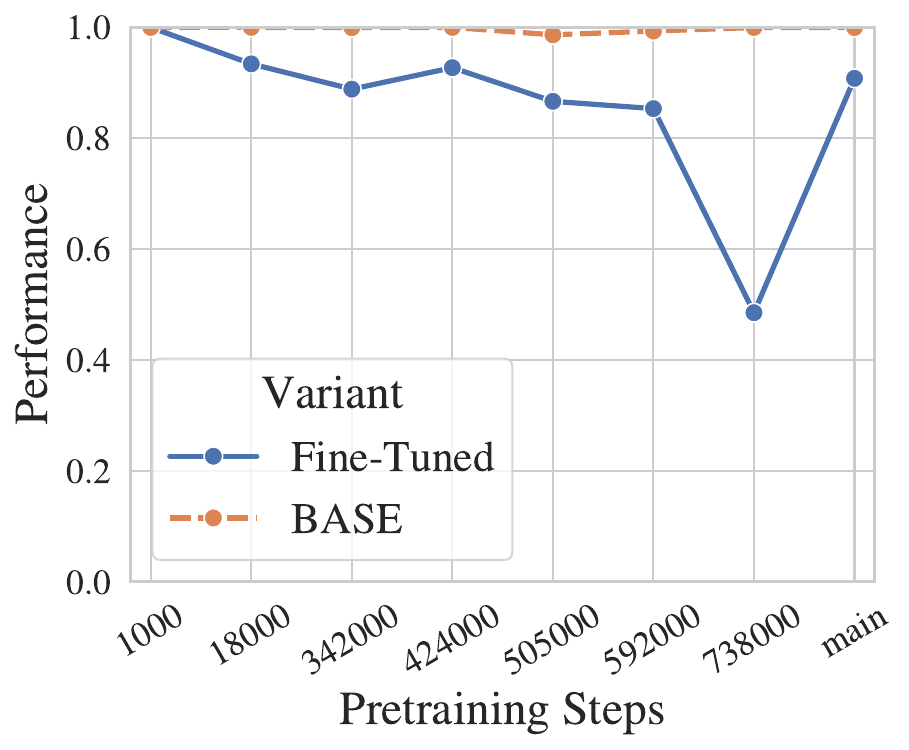}
        \caption{Paws -> QQP}
    \end{subfigure}%
    ~ 
    \begin{subfigure}[b]{0.25\textwidth}
    \includegraphics[width=\the\columnwidth]{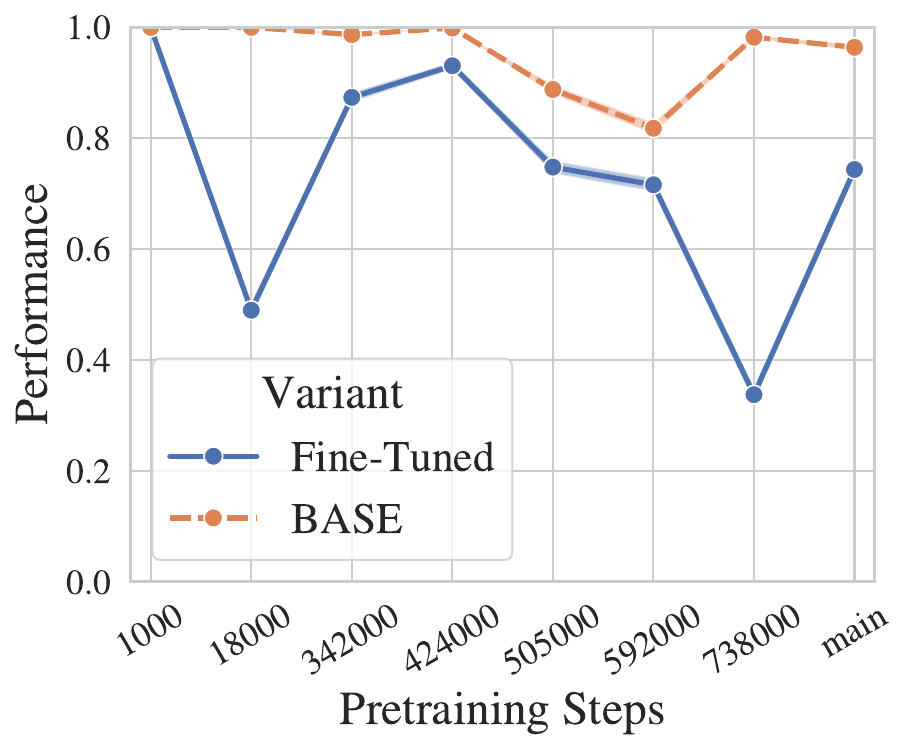}
        \caption{Paws -> STS-B}
    \end{subfigure}%
    ~ 
    \begin{subfigure}[b]{0.25\textwidth}
    \includegraphics[width=\the\columnwidth]{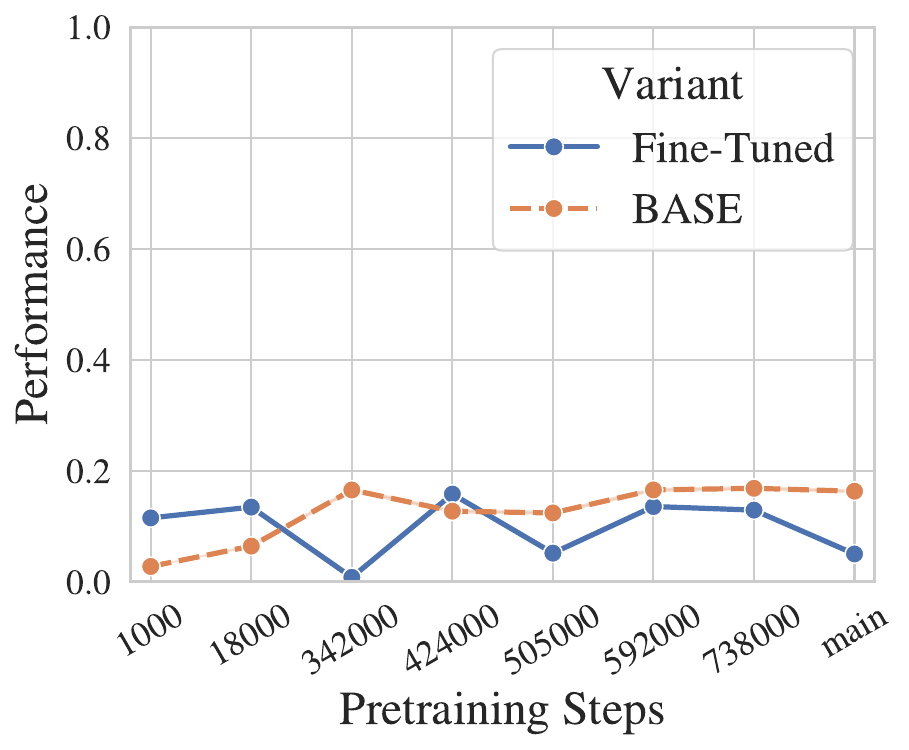}
        \caption{SocialIQA -> SciQ}
    \end{subfigure}%
    \begin{subfigure}[b]{0.25\textwidth}
    \includegraphics[width=\the\columnwidth]{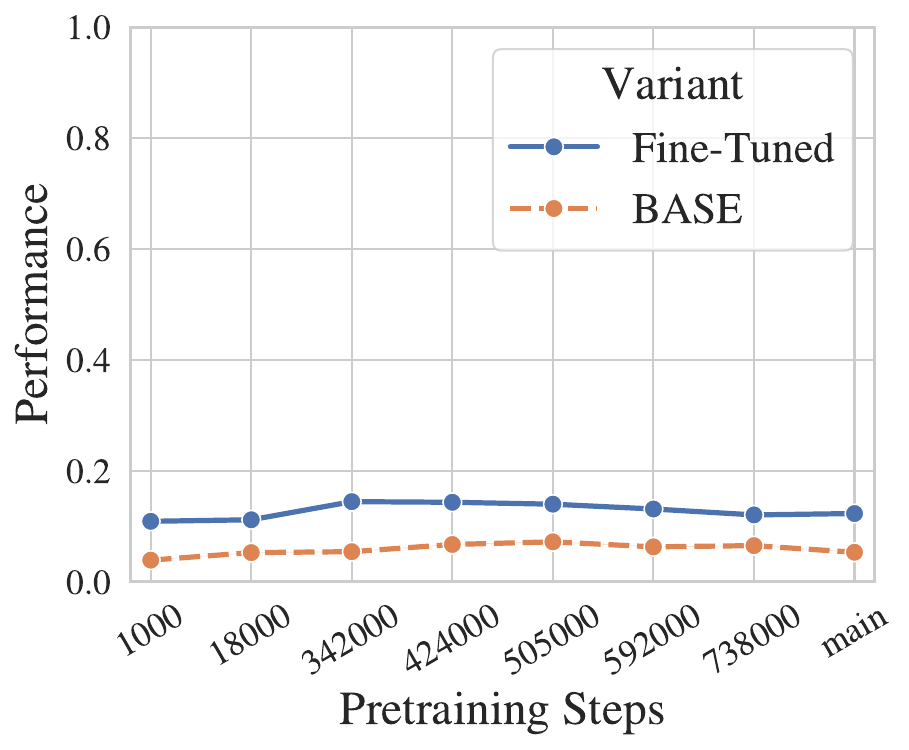}
        \caption{SocialIQA -> TweetQA}
    \end{subfigure}%
    \caption{Out-of-domain performance after supervised fine-tuning on each pre-training step.}
    \label{fig:ood-sft-ckpt-perf}
\end{figure*}

\textbf{Out-of-domain Generalization }
The out-of-domain performance for each dataset with respect to pre-training steps is shown in Figure~\ref{fig:ood-sft-ckpt-perf}.
Overall, the model generalizes well after fine-tuning on NLI tasks, while its performance deteriorates when evaluated on out-of-domain paraphrase detection tasks.

\begin{figure*}[t!]
    \centering
    \begin{subfigure}[b]{0.3\textwidth}
    \includegraphics[width=\the\columnwidth]{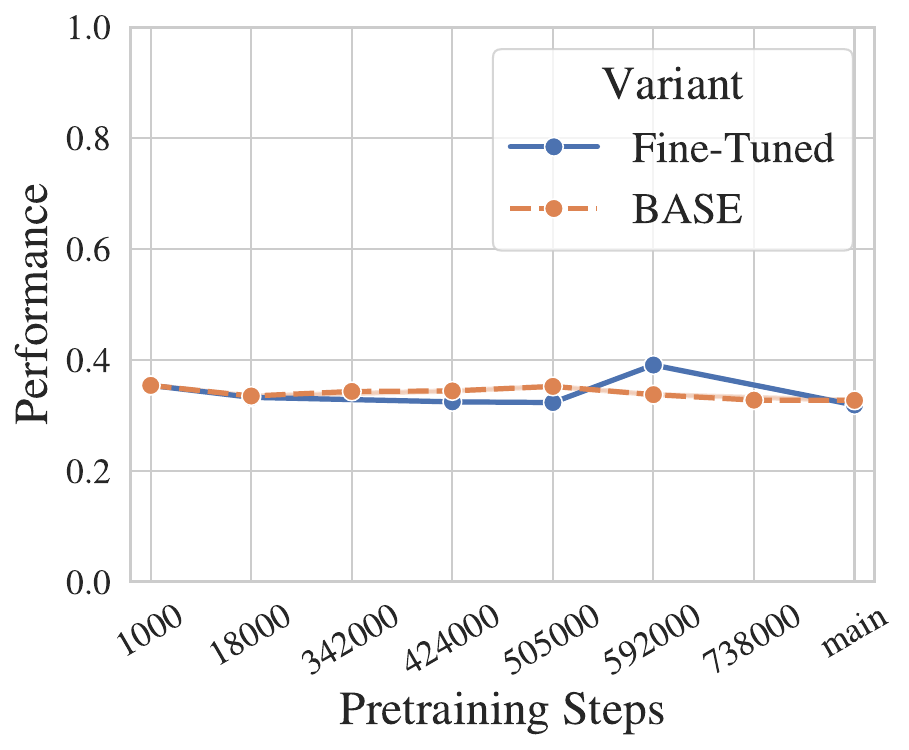}
        \caption{Paws -> MNLI}
    \end{subfigure}%
    ~ 
    \begin{subfigure}[b]{0.3\textwidth}
    \includegraphics[width=\the\columnwidth]{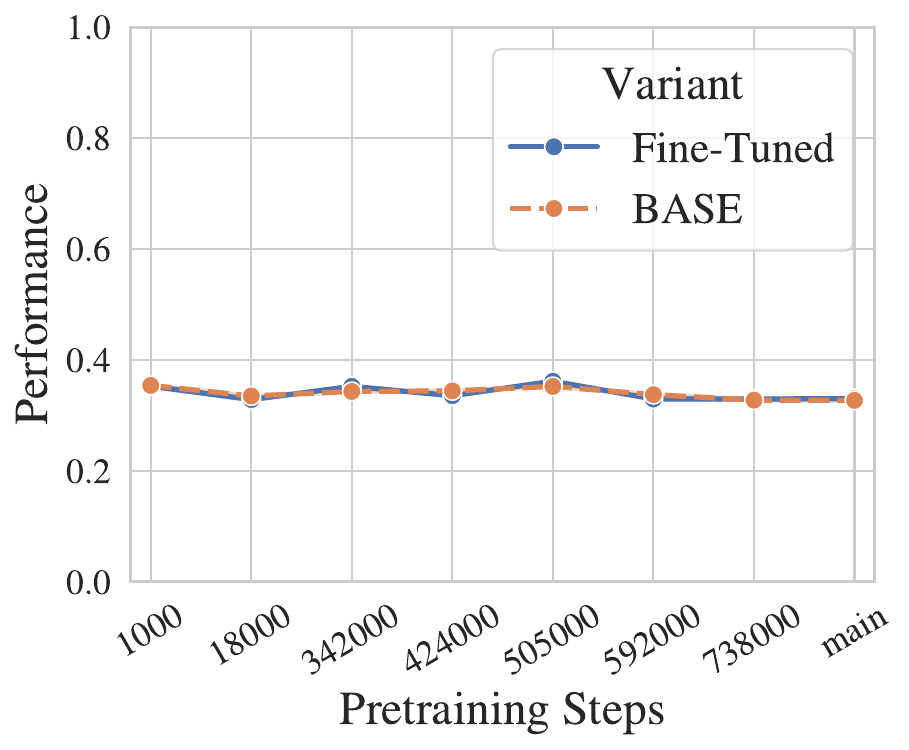}
        \caption{SocialIQA -> MNLI}
    \end{subfigure}%
    ~ 
    \begin{subfigure}[b]{0.3\textwidth}
    \includegraphics[width=\the\columnwidth]{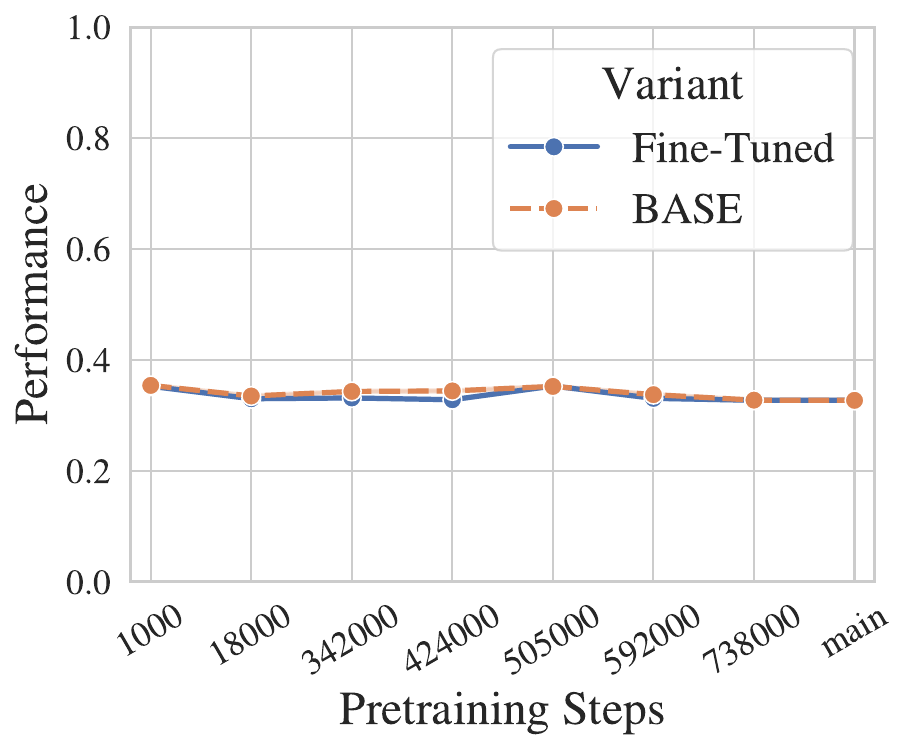}
        \caption{XSum -> MNLI}
    \end{subfigure}%
    \\
    \begin{subfigure}[b]{0.3\textwidth}
    \includegraphics[width=\the\columnwidth]{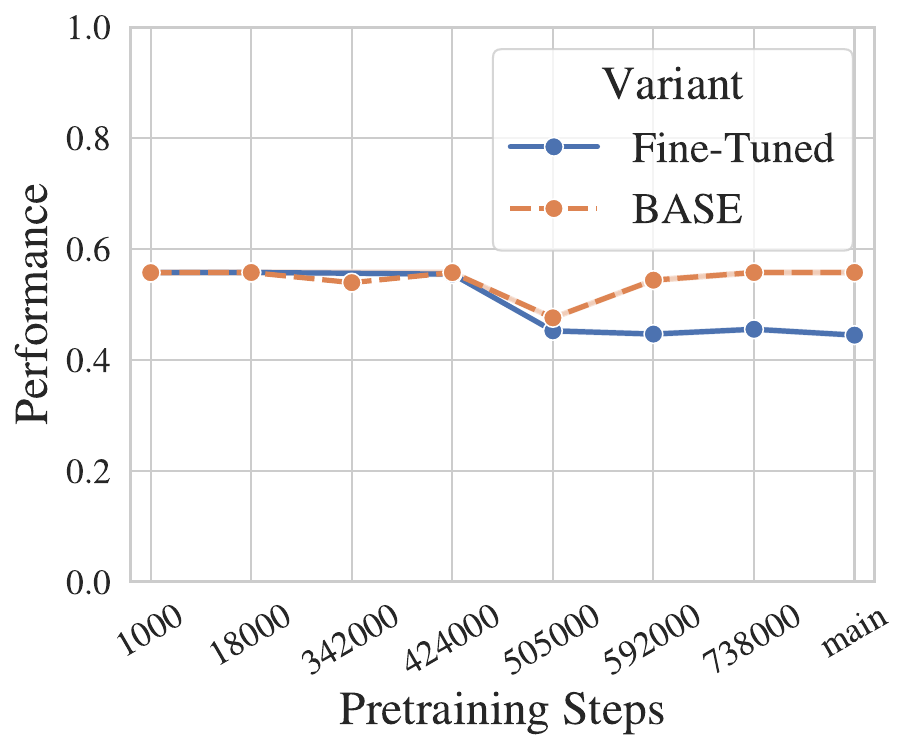}
        \caption{MNLI -> Paws}
    \end{subfigure}%
    ~ 
    \begin{subfigure}[b]{0.3\textwidth}
    \includegraphics[width=\the\columnwidth]{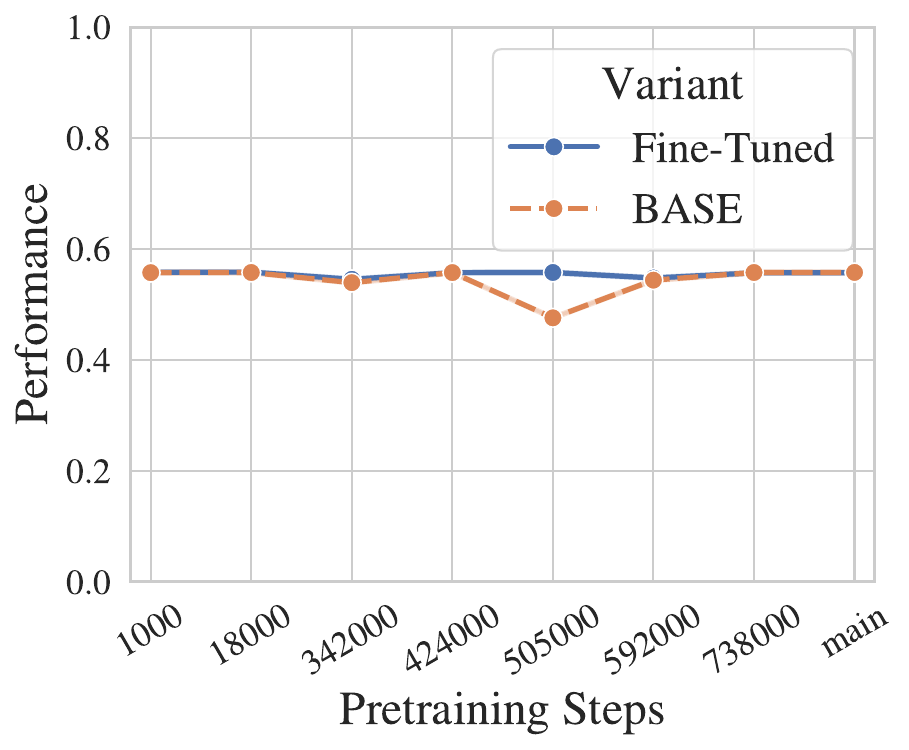}
        \caption{SocialIQA -> Paws}
    \end{subfigure}%
    ~ 
    \begin{subfigure}[b]{0.3\textwidth}
    \includegraphics[width=\the\columnwidth]{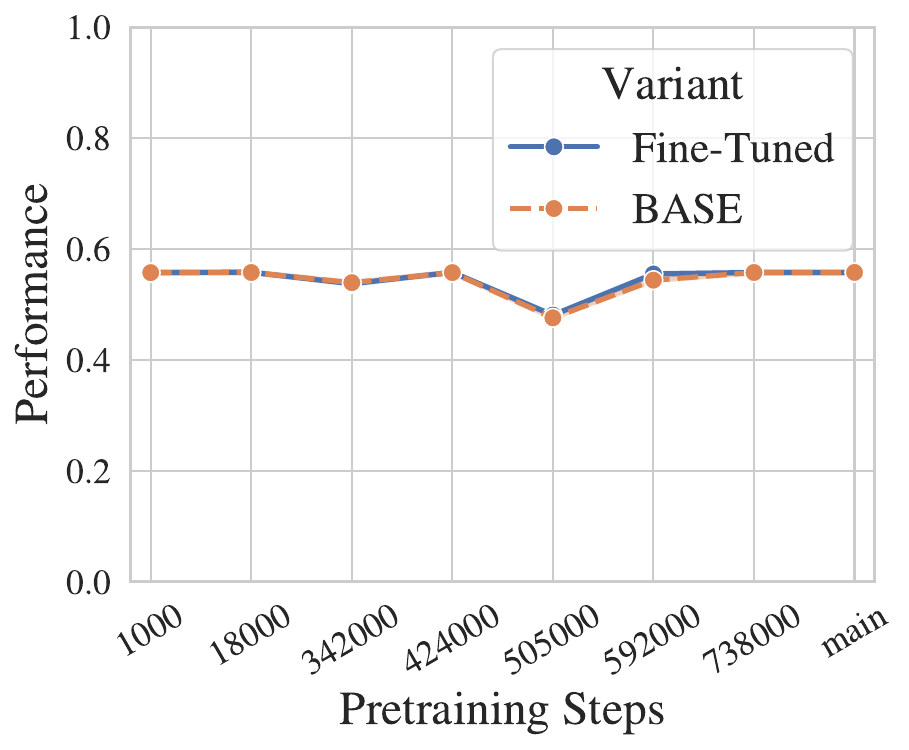}
        \caption{XSum -> Paws}
    \end{subfigure}%
    \\
    \caption{Cross-task performance after supervised fine-tuning on each pre-training step. The model is fine-tuned on a classification task and evaluated on a generation task or a classification task with a different label set.}
    \label{fig:cross-task-ckpt-perf-class}
\end{figure*}

\begin{figure*}[t!]
    \centering
    \begin{subfigure}[b]{0.3\textwidth}
    \includegraphics[width=\the\columnwidth]{figures/fig_files/cross-task/sft_evalpaws-trainxsum.pdf}
        \caption{Paws -> XSum}
    \end{subfigure}%
    ~ 
    \begin{subfigure}[b]{0.3\textwidth}
    \includegraphics[width=\the\columnwidth]{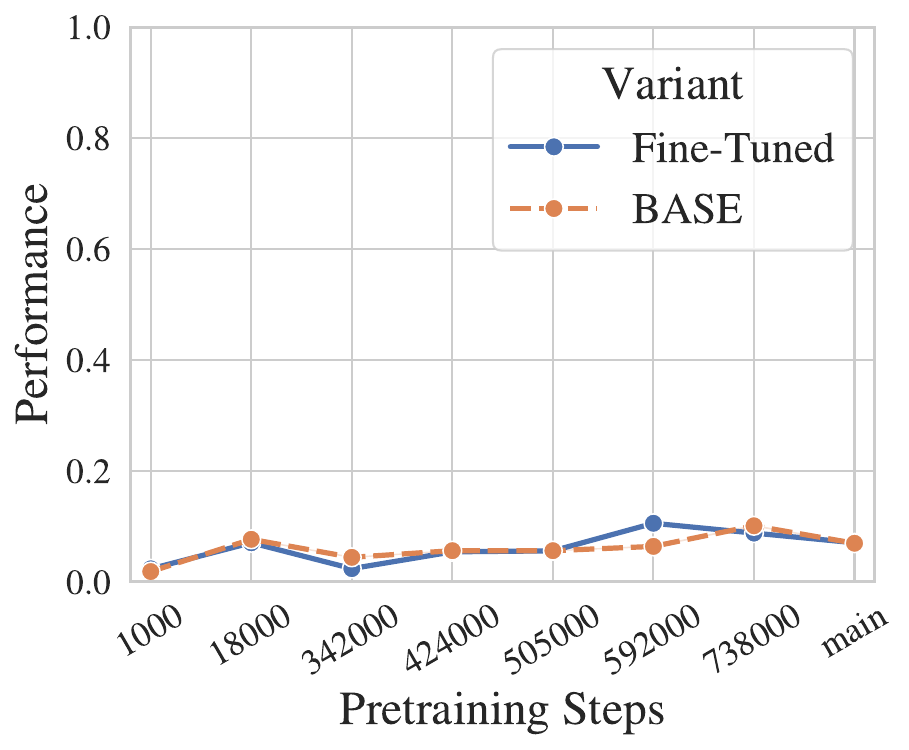}
        \caption{SocialIQA -> XSum}
    \end{subfigure}%
    ~ 
    \begin{subfigure}[b]{0.3\textwidth}
    \includegraphics[width=\the\columnwidth]{figures/fig_files/cross-task/sft_evalmnli_matched-trainxsum.pdf}
        \caption{MNLI -> XSum}
    \end{subfigure}%
    \\
    \begin{subfigure}[b]{0.3\textwidth}
    \includegraphics[width=\the\columnwidth]{figures/fig_files/cross-task/sft_evalpaws-trainsocialiqa.pdf}
        \caption{Paws -> SocialIQA}
    \end{subfigure}%
    ~ 
    \begin{subfigure}[b]{0.3\textwidth}
    \includegraphics[width=\the\columnwidth]{figures/fig_files/cross-task/sft_evalmnli_matched-trainsocialiqa.pdf}
        \caption{MNLI -> SocialIQA}
    \end{subfigure}%
    ~ 
    \begin{subfigure}[b]{0.3\textwidth}
    \includegraphics[width=\the\columnwidth]{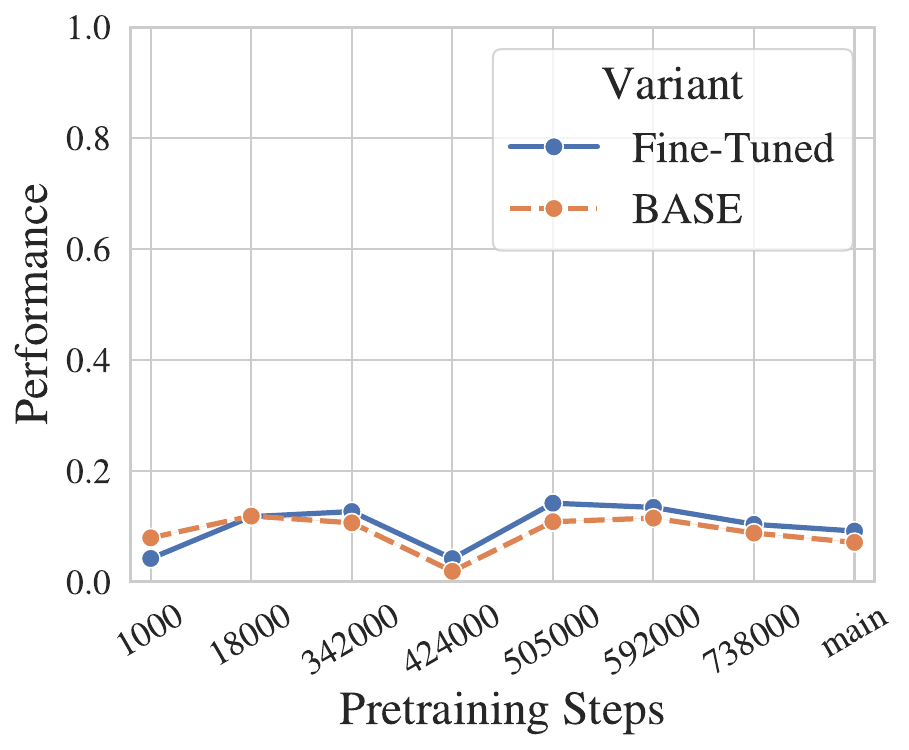}
        \caption{XSum -> SocialIQA}
    \end{subfigure}%
    \\
    \caption{Cross-task performance after supervised fine-tuning on each pre-training step. The model is fine-tuned on a generation task and evaluated on a classification task.}
    \label{fig:cross-task-ckpt-perf-gen}
    \vspace{-10pt}
\end{figure*}

\textbf{Cross-task Generalization }
The cross-task performance for each dataset with respect to pre-training steps is shown in Figure~\ref{fig:cross-task-ckpt-perf-class} and Figure~\ref{fig:cross-task-ckpt-perf-gen}.

\begin{figure*}[t!]
    \centering
    \begin{subfigure}[b]{0.4\textwidth}
    \includegraphics[width=\the\columnwidth]{figures/fig_files/task_format/task_format_evalmnli_matched-trainmnli.pdf}
        \caption{MNLI matched}
    \end{subfigure}%
    ~ 
    \begin{subfigure}[b]{0.4\textwidth}
    \includegraphics[width=\the\columnwidth]{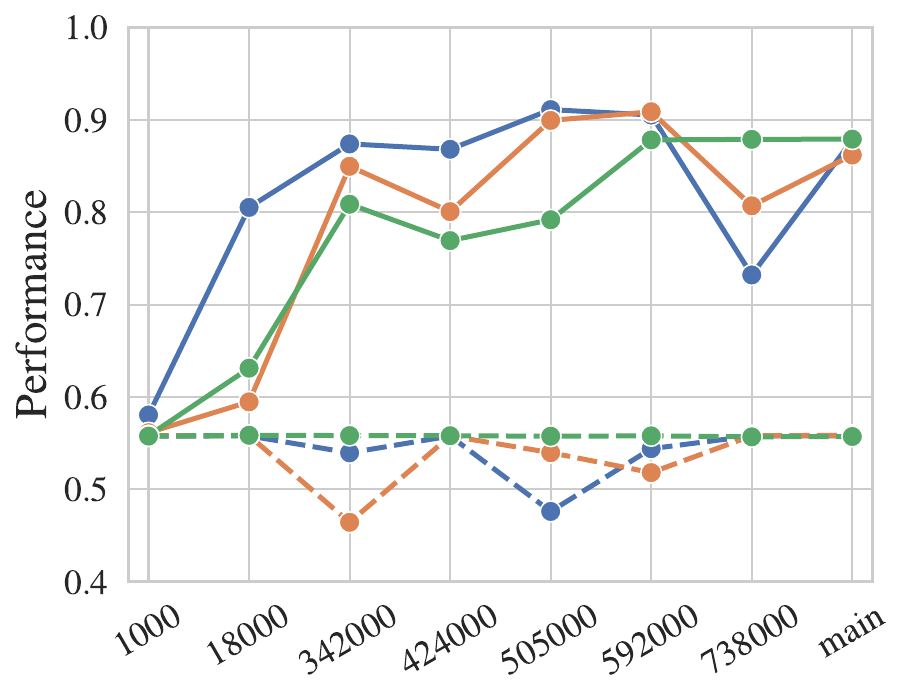}
        \caption{Paws}
    \end{subfigure}%
    \\
    \begin{subfigure}[b]{0.4\textwidth}
    \includegraphics[width=\the\columnwidth]{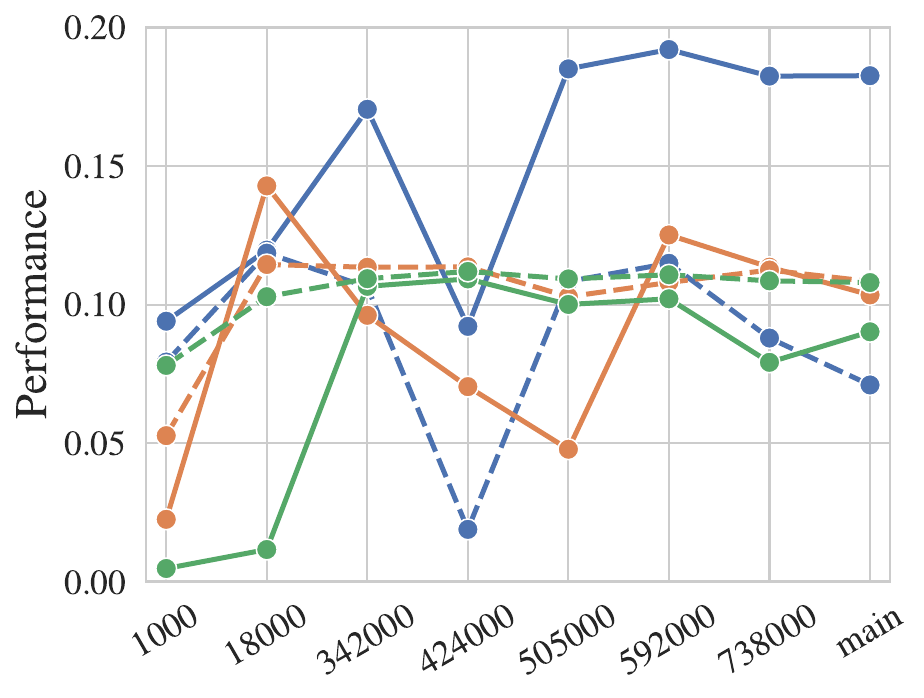}
        \caption{XSum}
    \end{subfigure}%
    ~ 
    \begin{subfigure}[b]{0.55\textwidth}
    \includegraphics[width=\the\columnwidth]{figures/fig_files/task_format/task_format_evalsocialiqa-trainsocialiqa.pdf}
        \caption{SocialIQa}
    \end{subfigure}%

    \caption {Model performance with different task formats. }
  \label{fig:app:task_format}
\end{figure*}

\textbf{Task-Format }
The performance of models on evaluation sets formatted with different prompt formatting methods is shown in Figure~\ref{fig:app:task_format}.

\begin{table*}[ht]
\scriptsize
\centering
\begin{tabular}{@{}cllll@{}}
\toprule
\multicolumn{1}{l}{\textbf{Task}} & \textbf{Default Prompt} & \textbf{Instruction Prompt} & \textbf{IO Prompt} & \textbf{Expected Output} \\ \midrule
\textbf{\begin{tabular}[c]{@{}c@{}}Summary\\ Generation\end{tabular}} & \begin{tabular}[c]{@{}l@{}}\#\#\# Input: \{document\}\\ \#\#\# Summary:\end{tabular} & \begin{tabular}[c]{@{}l@{}}Please read the following text: \{document\}\\ Provide a summary:\end{tabular} & \{document\} & \{summary\} \\ 
\arrayrulecolor{black!30}\midrule
\textbf{\begin{tabular}[c]{@{}c@{}}Question\\ Generation\end{tabular}} & \begin{tabular}[c]{@{}l@{}}\#\#\# Input: \{context\} \\ \#\#\# Answer: \{answer\}\\ \#\#\# Question:\end{tabular} & \begin{tabular}[c]{@{}l@{}}Given the context: \{context\}\\ And the answer: \{answer\}\\ Generate a suitable question:\end{tabular} & \begin{tabular}[c]{@{}l@{}}\{context\}\\ \{answer\}\end{tabular} & \{question\} \\ 
\arrayrulecolor{black!30}\midrule
\textbf{\begin{tabular}[c]{@{}c@{}}Natural Language\\ Inference\end{tabular}} & \begin{tabular}[c]{@{}l@{}}\#\#\# Input\_1: \{premise\} \\ \#\#\# Input\_2: \{hypothesis\}\\ \#\#\# Inference:\end{tabular} & \begin{tabular}[c]{@{}l@{}}Consider the following texts: Text 1: \{premise\} \\ Text 2: \{hypothesis\} The relation is\end{tabular} & \begin{tabular}[c]{@{}l@{}}\{premise\}\\ \{hypothesis\}\end{tabular} & \{label\} \\
\arrayrulecolor{black!30}\midrule
\textbf{Paraphrase Detection} & \begin{tabular}[c]{@{}l@{}}\#\#\# Input\_1: \{sentence1\} \\ \#\#\# Input\_2: \{sentence2\}\\ \#\#\# Paraphrase Classification:\end{tabular} & \begin{tabular}[c]{@{}l@{}}Let's compare the two sentences: \\ Sentence\_1: \{sentence1\}\\ Sentence\_2: \{sentence2\} Are they paraphrasing?:\end{tabular} & \begin{tabular}[c]{@{}l@{}}\{sentence1\}\\ \{sentence2\}\end{tabular} & \{label\} \\ \arrayrulecolor{black}\bottomrule
\end{tabular}
\caption{Formatting of the prompts}
\label{tab:app:promptformat}
\vspace{-20pt}
\end{table*}

\section{Llama3-8B Results}
\label{app:sec:llama}
\begin{table}[ht]
\centering
\small
\begin{tabular}{@{}rrr@{}}
\toprule
\multicolumn{1}{l}{\textbf{Checkpoint }} & \multicolumn{1}{c}{\begin{tabular}[c]{@{}c@{}}\textbf{  Learned in  } \\ \textbf{  Pre-train  }\end{tabular}} & \multicolumn{1}{c}{\begin{tabular}[c]{@{}c@{}}\textbf{  Learned in  } \\ \textbf{  Fine-Tune  }\end{tabular}} \\ \midrule
1000 & 0.048 & 0.062 \\
18000 & 0.048 & 0.149 \\
342000 & 0.004 & 0.286 \\
424000 & 0.01 & 0.297 \\
505000 & 0.03 & 0.304 \\
592000 & 0.027 & 0.297 \\
738000 & 0.021 & 0.264 \\
main & -0.005 & 0.290 \\ \bottomrule
\end{tabular}
\caption{Average performance change before and after fine-tuning for each checkpoint (Perf(Fine-tuned) - Perf(BASE)). The group that is never learned during pre-training is picked up by the model during fine-tuning.}
\label{tab:gain-tab-by-step}
\end{table}

\begin{figure}[ht]
\centering
  \includegraphics[width=0.8\columnwidth]{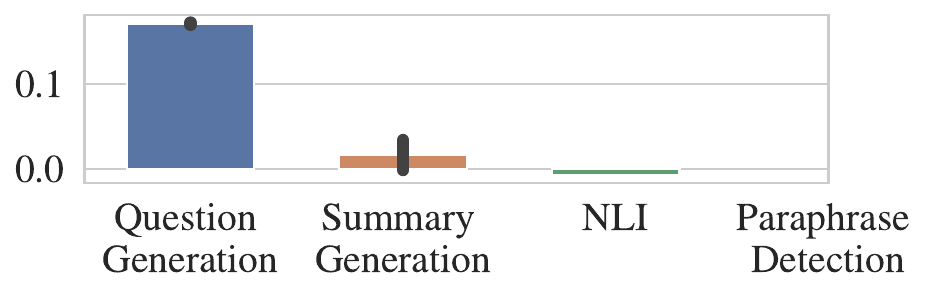}
  \caption{Ratio of out-of-domain performance change for each task on the final checkpoint of LLAMA3-8B.}
  \label{fig:finding:ood-by-data-llama}
\end{figure}

To provide more evidence of the findings on a different model architecture and size, we conduct some experiments on the final checkpoint of Llama3-8B.

\textbf{Task Transfer } 
Similar to our findings with OLMo, Llama3-8B fine-tuned on classification tasks and evaluated on generation tasks decreases on average 61.0\% compared to models that are never fine-tuned. 
In contrast, models fine-tuned on generation tasks perform similarly to the BASE model on classification tasks, with a 10.6\% MRC.

\textbf{Domain Knowledge }
The ratio of out-of-domain performance change by task is shown in Figure~\ref{fig:finding:ood-by-data-llama}.
Overall, we observe that Llama and OLMo experience benefits with different tasks after fine-tuning, but both model shows an inconsistent change across tasks.

\begin{table*}[ht]
\scriptsize
\centering
\begin{tabular}{ccc|cc}
\toprule
\textbf{Name} & \textbf{License} &  & \textbf{Name} & \textbf{License} \\ \midrule
OLMo-1b & Apache 2.0 &  & SocialIQa & CC-BY \\
TULU & ODC-BY &  & \text{  }CNN/DailyMail & Apache 2.0 \\
ARC & CC BY-SA &  & TweetQA & CC BY-SA-4.0 \\
OpenbookQA & Apache 2.0 &  & MNLI & CC-BY-3.0 \\
Hellaswag & MIT &  & GPT3NLI & MIT \\
BoolQ & Apache 2.0 &  & RTE & N/A \\
SciQ & CC-BY-NC-3.0 &  & Paws & Free \\
XSum & MIT &  & QQP & Non-Commercial \\
XLSum & CC-BY-NC-SA 4.0 \text{  } &  & STS-B & Other \\ \bottomrule
\end{tabular}
\caption{License of artifacts used in this paper.}
\label{app:tab:artifact}
\end{table*}

\section{License of Artifacts}
We include the license of artifacts used in this paper in Table~\ref{app:tab:artifact}

\section{Full Performance Table}
Due to the availability of space and the amount of fine-tuned checkpoints, we omit displaying all exact metric values in the paper.
The performance of each fine-tuned variant on each dataset can be found in the \texttt{csv} file under directory \texttt{results} in the code base.

\section{Performance Difference Numbers}
\label{sec:app:performance-numbers}

The average performance change before and after fine-tuning for each checkpoint is shown in Table~\ref{tab:gain-tab-by-step}.
The data in this table is used to create Figure~\ref{fig:finding:ptftcompare}.

\section{Generalization Taxonomy}
\label{sec:app:generalization-taxo}
Following the generalization taxonomy in \citet{hupkes2023taxonomy}, the evaluation card is included in Table~\ref{app:tab:gen_eval_card}.
\newcommand{\tabularwidth}{\textwidth}
\newcommand{\expone}{$\square$}
\newcommand{\exptwo}{$\bigtriangleup$}

\begin{table*}[h] %
\centering
\renewcommand{\arraystretch}{1.1}         
\setlength{\tabcolsep}{0mm}         

\begin{tabular}{|p{\tabularwidth}<{\centering}|}         
\hline

\rowcolor{gray!60}\textbf{Motivation} \\               
\footnotesize
\begin{tabular}{p{0.25\tabularwidth}<{\centering} p{0.25\tabularwidth}<{\centering} p{0.25\tabularwidth}<{\centering} p{0.25\tabularwidth}<{\centering}}                        
\textit{Practical} & \textit{Cognitive} & \textit{Intrinsic} & \textit{Fairness}\\
\expone\hspace{0.8mm}\exptwo\hspace{0.8mm}
& 
& 
& 

\vspace{2mm} \\
\end{tabular}\\

\rowcolor{gray!60}\textbf{Generalisation type} \\               
\footnotesize
\begin{tabular}{m{0.21\tabularwidth}<{\centering} m{0.2\tabularwidth}<{\centering} m{0.13\tabularwidth}<{\centering} m{0.13\tabularwidth}<{\centering} m{0.13\tabularwidth}<{\centering} m{0.2\tabularwidth}<{\centering}}                   
\textit{Compositional} & \textit{Structural} & \textit{Cross Task} & \textit{Cross Language} & \textit{Cross Domain} & \textit{Robustness}\\
& 
& \exptwo\hspace{0.8mm}
& 
& \expone\hspace{0.8mm}
& 

\vspace{2mm} \\
\end{tabular}\\

\rowcolor{gray!60}\textbf{Shift type} \\             
\footnotesize
\begin{tabular}{p{0.25\tabularwidth}<{\centering} p{0.25\tabularwidth}<{\centering} p{0.25\tabularwidth}<{\centering} p{0.25\tabularwidth}<{\centering}}                        
\textit{Covariate} & \textit{Label} & \textit{Full} & \textit{Assumed}\\  
\expone\hspace{0.8mm}
& \exptwo\hspace{0.8mm}
&
& 

\vspace{2mm} \\
\end{tabular}\\

\rowcolor{gray!60}\textbf{Shift source} \\             
\footnotesize
\begin{tabular}{p{0.25\tabularwidth}<{\centering} p{0.25\tabularwidth}<{\centering} p{0.25\tabularwidth}<{\centering} p{0.25\tabularwidth}<{\centering}}                          
\textit{Naturally occuring} & \textit{Partitioned natural} & \textit{Generated shift} & \textit{Fully generated}\\
\expone\hspace{0.8mm}\exptwo\hspace{0.8mm}
&
&
& 

\vspace{2mm} \\
\end{tabular}\\

\rowcolor{gray!60}\textbf{Shift locus}\\             
\footnotesize
\begin{tabular}{p{0.25\tabularwidth}<{\centering} p{0.25\tabularwidth}<{\centering} p{0.25\tabularwidth}<{\centering} p{0.25\tabularwidth}<{\centering}}                         
\textit{Train--test} & \textit{Finetune train--test} & \textit{Pretrain--train} & \textit{Pretrain--test}\\
&
\expone\hspace{0.8mm}\exptwo\hspace{0.8mm}
&
&
\vspace{2mm} \\
\end{tabular}\\

\hline
\end{tabular}

\caption{Generalization experiment summary following taxonomy in \citet{hupkes2023taxonomy}.}
\label{app:tab:gen_eval_card}
\end{table*}

\label{sec:appendix}

\end{document}